\documentclass[twocolumn]{article} 

\usepackage{amsmath}
\usepackage{mathtools}
\usepackage{amssymb}
\usepackage{mathabx}        
\usepackage{bbold}          
\usepackage{bm}
\usepackage{soul}
\usepackage{xcolor}

\usepackage{enumitem}
\usepackage{url}
\usepackage{comment}
\usepackage{subfigure}
\usepackage{tabularx}
\usepackage{booktabs}
\usepackage[hidelinks]{hyperref}
\usepackage{cite}

\usepackage{tikz}
\usetikzlibrary{calc,angles,positioning,intersections,quotes,decorations.markings}

\usepackage{amsthm}
\usepackage{geometry}
\newcommand{\RN}{\mathbb{R}} 				

\newcommand{\BR}[1]{\left({ #1 }\right)}	
\newcommand{\CBR}[1]{\left\lbrace #1 \right\rbrace}
\newcommand{\norm}[1]{\left\Vert #1\right\Vert}	

\newcommand{\PARDIFF}[2]{\ensuremath{\frac{\partial #1}{\partial #2}}}

\newcommand{\DET}[1]{\textbf{det}\BR{ #1 }}

\newtheorem{theorem}{Theorem}
\newtheorem{lemma}{Lemma}
\newtheorem{proposition}{Proposition}
\newtheorem{corollary}{Corollary}

\title{\Large{ \textbf{Securing Isosceles Triangular Formations under Heterogeneous Sensing and Mixed Constraints}}}
\author{Nelson P.K. Chan, Bayu Jayawardhana, Hector Garcia de Marina
\thanks{Nelson and Bayu are with Engineering and Technology Institute Groningen, 
    University of Groningen, 
    Nijenborgh 4, 9747AG,
    Groningen, the Netherlands. 
    \newline
	The work of N.P.K. Chan \& B. Jayawardhana is supported by the Region of Smart Factories (ROSF) project financed by REP-SNN and by the STW Smart Industry 2016 programme.
    }
\thanks{Hector is with Department of Computer Architecture and Automatic Control at the Faculty of Physics, 
    Universidad Complutense de Madrid,
    28040, Madrid, Spain.
    \newline
	The work of H.G. de Marina is supported by the grant \emph{Atraccion de Talento} 2019-T2/TIC-13503 from the Government of the Autonomous Community of Madrid.
    }
}
\date{ }

\begin{document}

\maketitle

\begin{abstract}
This paper focuses on securing a triangular shape (up to translation) for a team of three mobile robots that uses heterogeneous sensing mechanism.
Based on the available local information, each robot employs the popular gradient-based control law to attain the assigned individual task(s).
In the current work, robots are assigned either distance and signed area task(s) or bearing task(s). 
We provide a sufficient condition on the gain ratio $ R_{\text{Ad}} $ between the signed area and the distance control term such that the desired formation shape, an isosceles triangle, is reached from all feasible starting positions. 
Numerical simulations are provided to support the theoretical analyses.
\end{abstract}

\section{Introduction} \label{sec:Introduction}
In the presence of limited sensing range, certain complex tasks, such as covering a region of interest, can be executed within a smaller amount of time by letting a team of robots move in formation than when it is carried out by a single robot. 
In addition, each robot within the formation can be equipped with different sensors and therefore, more (local) information can be gathered \cite{Anderson2008}. 
With this application perspective, the study on attaining and maintaining a particular geometric shape by a team of robots has attracted  research interests; see \cite{Oh2015, Zhao2019, Chen2020} for an overview of approaches whereby the desired formation shape is described by a set of only inter-robot relative position, distance, bearing, or angle constraints. 

Recently, approaches on realizing a formation shape using a mixed set of constraints have been considered.  
In \cite{Kwon2018}, distance and angle constraints are employed; the paper \cite{Bishop2013} considers the combination of distance and bearing constraints, while the work in  \cite{Kwon2019a} use distance, bearing, and angle constraints to describe the desired formation shape.
Furthermore, a series of works \cite{Anderson2017a, Sugie2018, Cao2019a, Sugie2020} focus on the combination of distance and signed area constraints.
By adding signed area constraints, the authors tackle the flip and flex ambiguity problem present in distance-based formation control. 

Except for \cite{Sugie2018, Cao2019a, Sugie2020}, the referred works deal with an undirected graph.
Moreover, it is usually assumed that all robots within the formation sense a common (geometric) variable, such as a relative position, distance, or bearing.

In formation control, triangular formations consisting of three autonomous agents serve as a class of benchmarks that can be used to test and compare the performances of different controllers. 
This apparently simple setup allows detailed rigorous analysis for novel techniques and methodologies as we aim at in this work (see \cite{Cao2007, anderson2007control, Marina2017, Cao2008, liu2014controlling, 7039453}), and therefore, it provides a starting point in order to achieve more general formations.

Our earlier work \cite{Chan2020} dealt with the formation control problem in which the robots can have different sensing mechanism and as a result also different individual task(s).
In particular, we partitioned the team of robots in two categories, namely distance robots carrying out distance tasks and bearing robots fulfilling bearing constraints.
We mention that the interconnection topology is a directed graph. 
For the particular case of one distance and two bearing robots, the (\textbf{1D2B}) setup, we showed the existence of moving configurations, i.e., where the robots converge to a non-zero translation velocity, which are locally attractive under certain conditions. 
One observation on the moving configurations is that the signed area of the corresponding shape has an opposite sign when compared with the desired formation shape, i.e., the robots converge to an \textit{incorrect} shape. 

Building on \cite{Chan2020}, in this work, we aim to avoid the occurrence of moving configurations by adding a signed area constraint to the distance robot. 
We remark that this does not increase the sensing load on the robot. 
For the current analysis, we consider isosceles triangles as the desired formation shape. 
Our main contribution is then the identification of a specific constraint on the gain ratio $ R_{\text{Ad}} $ between the signed area and the distance control term yielding robots to always evolve to the correct isosceles triangular shape.

The remainder of this paper is organized as follows: 
We first present some preliminaries in Section \ref{sec:Preliminaries}.
The (\textbf{1D2B}) setup with the added signed area constraint is then considered in Section \ref{sec:Problem-Setup}.
We provide preliminary analysis for the specific case of isosceles triangles in Section \ref{sec:Isosceles-Triangle-Analysis}.
In Sections \ref{sec:Isosceles-Triangle-Analysis-Acute} and \ref{sec:Isosceles-Triangle-Analysis-Right-Obtuse}, we then consider more in depth the case of acute isosceles triangles and right and obtuse isosceles triangles, respectively. 
The theoretical claims are supported by numerical results in Section \ref{sec:Numerical-Example}. 
Finally, we end with the conclusion in Section \ref{sec:Conclusions}.

\paragraph*{Notation}
For a vector $ x \in \RN^{n} $, $ x^{\top} $ is the transpose, and $ \norm{x} = \sqrt{x^{\top}x} $ the $ 2 $-norm.
The vector $ \mathbb{1}_{n} $ (or $ \mathbb{0}_{n} $) denotes the vector with entries being all $ 1 $s (or $ 0 $s).
In the plane $ \RN^{2} $, the symbol $ \angle v_{1} $ denotes the counter-clockwise angle from the $ x $-axis of a coordinate frame $ \Sigma $ to the vector $ v_{1} \in \RN^{2} $. 
The matrix $ J = \left[ \begin{smallmatrix} 0 & 1 \\ -1 & 0 \end{smallmatrix} \right] $ is the rotation matrix with angle $ -90^{\degree} $. 
We denote $ x^{\perp} $ as the perpendicular vector obtained by rotating $ x $ with a counter-clockwise angle of $ +90^{\degree} $; we have $ x^{\perp} = - J x $.
For two points $ p_{i} $ and $ p_{j} $ in the plane, we define the relative position vector as $ z_{ij} = p_{j} - p_{i} \in \RN^{2} $, the distance as $ d_{ij} = \norm{z_{ij}} \in \RN_{\geq 0} $ and the relative bearing vector, when $d_{ij}\neq 0, $ as $ g_{ij} = \frac{z_{ij}}{d_{ij}} \in \RN^{2} $, all relative to a global coordinate frame $ \Sigma^{\text{g}} $. 
It follows $ z_{ji} = - z_{ij} $, $ d_{ji} = d_{ij} $ and $ g_{ji} = - g_{ij} $.

\section{Preliminaries} \label{sec:Preliminaries}

\subsection{Robot configurations} \label{subsec:Robot-Conf}
We consider a team consisting of three robots in which \texttt{Ri} is the label assigned to robot $ i $.
The robots are moving in the plane $ \RN^{2} $ according to the single integrator dynamics, i.e., 
\begin{equation} \label{eq:Prel-Robot-Dynamics}
    \begin{aligned}
        \dot{p}_{i}\BR{t} = u_{i}\BR{t}, \quad i \in \CBR{1, \, 2, \, 3}
        ,
    \end{aligned}
\end{equation}
where $ p_{i} \in \RN^{2} $ (a point in the plane) and $ u_{i} \in \RN^{2} $ represent the position of and the control input for \texttt{Ri}, respectively.
For convenience, all spatial variables are given relative to a global coordinate frame $ \Sigma^{\text{g}} $.
We assume $ p_{i}\BR{t^{*}} \neq p_{j}\BR{t^{*}}, \, \forall t^{*} \in \left[0, \, t \right) $ if $ i \neq j $, i.e., two robots cannot occupy the same position at the same time. 

The vector $ p = \begin{bmatrix} p_{1}^{\top} & p_{2}^{\top} & p_{3}^{\top} \end{bmatrix}^{\top} \in \RN^{6} $ represents the \textit{team configuration}.
We define $ p_{\text{ref}} $  as the \textit{reference configuration} where the three positions describe a particular triangle $ T $ of interest up to translation. 
Therefore, the set of \textit{desired configurations} can be formally written from the reference configuration $ p_{\text{ref}} $ as
\begin{equation} \label{eq:Prel-Desired-Configurations-p}
    \mathcal{S}_{p} := \CBR{p \in \RN^{6} \, \rvert \, p = p_{\text{ref}} + \BR{\mathbb{1}_{3} \otimes v}, v\in\RN^2}
    , 
\end{equation}
where $ v$ is a translation vector and $ \otimes $ denotes the Kronecker product. 
Note that the desired configurations can also be described in terms of relative positions (or links) $ z_{ij} $'s with $ z_{ij} = p_{j} - p_{i} \in \RN^{2} $.
In particular, it is the singleton $ \mathcal{S}_{z} $ whose one element is
\begin{equation} \label{eq:Prel-Desired-Configurations-z}
    z_{\text{ref}} = \BR{H \otimes I_{2}} p_{\text{ref}},
    \quad 
    H = 
    \begin{bmatrix} 
        -1 & 1 & 0 
        \\ 
        -1 & 0 & 1 
        \\ 
        0 & -1 & 1 
    \end{bmatrix}
\end{equation}
where $ I_{2} $ is the identity matrix.

Consider the link vector $ z\BR{t} := \BR{H \otimes I_{2}} p\BR{t} $ converging to a point $ \widetilde{z} \in \RN^{6} $ and $ \dot{p}\BR{t} \to (\mathbb{1}_{3} \otimes w) $ as time progresses, where $ w \in \RN^{2} $ denotes a constant velocity vector. 
Let $ \widetilde{p} \in \RN^{6} $ denotes a configuration yielding $ \widetilde{z} $. 
Then, the team trajectory converges to $ p_{\text{traj}}\BR{t} = \BR{\mathbb{1}_{3}  \otimes c_{0}} + \widetilde{p} + \BR{\mathbb{1}_{3} \otimes w} t $, where $c_{0} \in \RN^2 $ is an arbitrary offset given by the initial condition $ p\BR{0} $. We have an \textit{equilibrium configuration} when $ w = \mathbb{0}_{2} $, and otherwise, the configuration is \textit{moving}.
In the latter case, $ p_{\text{traj}}\BR{t} \to \infty $ as $ t \to \infty $.
Depending on whether $ \widetilde{z} \in \mathcal{S}_{z} $ or not, we classify the obtained configuration $ \widetilde{p} $ as \textit{desired} or \textit{incorrect}.

\subsection{Signed area of a triangle} \label{subsec:Signed-Area}
For a triangle $ T $ with points $ p_{1} $, $ p_{2} $, and $ p_{3} $ in the plane $ \RN^{2} $, the \textit{signed area} $ A $ is given by 
\begin{equation} \label{eq:Prel-Signed-Area}
    \begin{aligned}
        A
            = \frac{1}{2} \DET{
            \begin{bmatrix}
                1 & 1 & 1
                \\
                p_{1} & p_{2} & p_{3}
            \end{bmatrix}
            },
    \end{aligned}
\end{equation}
where $ \DET{\bullet} $ denotes the determinant of a matrix.
Alternative expressions for $ A $ are $ A = \frac{1}{2} z_{12}^{\top} J z_{13} $ or $ A = \frac{1}{2} \sin \theta_{213} \, d_{12} d_{13} $ with $ \sin \theta_{213} = g_{12}^{\top} J g_{13} $ and $ \theta_{213} $ being the \textit{signed angle} enclosed by the bearing vectors $ g_{12} $ and $ g_{13} $ at point $ p_{1} $. 
Assuming $ \angle g_{12} = \alpha $ and $ \angle g_{13} = \beta $ with respect to a coordinate frame $ \Sigma $, it follows that $ \sin \theta_{213} = \sin \BR{\beta - \alpha} $.
Hence for a counter-clockwise (or clockwise) ordering of points $ p_{1} $, $ p_{2} $, and $ p_{3} $, we obtain $ \theta_{213} > 0 $ and also $ A > 0 $ (or $ \theta_{213} < 0 $ and also $ A < 0 $). 

\subsection{Cubic equations} \label{subsec:Cubic-Equations}

Consider a \textit{reduced}\footnote{in some texts, the term `depressed' is used.} cubic equation
\begin{equation} \label{eq:Prel-Reduced-Cubic}
    y^{3} + cy + d = 0
    ,
\end{equation}
where $ c, \, d \in \RN $ and discriminant $ \Delta = -4 c^{3} - 27 d^{2} $.

\begin{lemma}[\cite{Chan2020}] \label{lem:Prel-Reduced-Cubic-Roots-Positive} 
    Given a reduced cubic equation \eqref{eq:Prel-Reduced-Cubic} with coefficients $ c < 0 $ and $ d > 0 $. 
    Assume the discriminant is $ \Delta \geq 0 $. 
    Then two positive real roots exist with values
    \begin{equation} \label{eq:Prel-Reduced-Cubic-Roots-Positive}
        \begin{aligned}
            y_{\text{p}_{1}} 
                & = 2 \sqrt[3]{r_{v}} \cos \BR{\frac{1}{3} \varphi_{v} - 120^{\degree}}
                & & \in \left(0, \, 1\right] \sqrt[3]{r_{v}}
                ,
                \\
            y_{\text{p}_{2}} 
                & = 2 \sqrt[3]{r_{v}} \cos \BR{\frac{1}{3} \varphi_{v}}
                & & \in \left[1, \, \sqrt{3}\right) \sqrt[3]{r_{v}}
            ,
        \end{aligned}
    \end{equation} 
    where $ r_{v} = \sqrt{- \BR{\frac{c}{3}}^{3}} $ and $ \varphi_{v} = \tan^{-1} \BR{\frac{-2}{d}\sqrt{\frac{\Delta}{108}}} \in \left(90^{\degree}, \, 180^{\degree} \right] $.
    When $ \Delta = 0 $, the two positive real roots are equal and have value $ y_{\text{p}_{1}} = y_{\text{p}_{2}} = \sqrt[3]{r_{v}} = \sqrt[3]{\frac{d}{2}} $.
\end{lemma}

\section{The (\textbf{1D2B}) Setup with a Signed Area Constraint} \label{sec:Problem-Setup}
As discussed in the Introduction, we consider the setup in which the robots possess heterogeneous sensing mechanism; as a result, they each fulfill different task(s) within the formation. 
Without loss of generality, we let robot \texttt{R1} be the \textit{distance robot} and \texttt{R2} and \texttt{R3} be \textit{bearing robots}.
Since \texttt{R1} is the distance robot, it needs to maintain \textit{distance constraint} $ d_{12}^{\star} $ with \texttt{R2} and similarly, $ d_{13}^{\star} $ with \texttt{R3}. 
It possesses an \textit{independent local coordinate frame} $ \Sigma^{\text{1}} $ which is \textit{not necessarily aligned} with that of \texttt{R2} and \texttt{R3} or the global coordinate frame $ \Sigma^{\text{g}} $.
Within $ \Sigma^{\text{1}} $, \texttt{R1} obtains the relative position information $ z_{12} $ and $ z_{13} $ from the neighboring robots. 
Robot \texttt{R2} has the task to keep a \textit{bearing constraint} $ g_{21}^{\star} $ relative to \texttt{R1}. 
The local coordinate frame $ \Sigma^{2} $ is \textit{aligned with} $ \Sigma^{\text{g}} $ and \texttt{R2} obtains the bearing measurement $ g_{21} $ within $ \Sigma^{2} $.
In a similar fashion, \texttt{R3} obtains bearing measurement $ g_{31} $ with respect to $ \Sigma^{3} $ and is required to satisfy a desired bearing $ g_{31}^{\star} $ relative to \texttt{R1}. 
In our previous work \cite{Chan2020}, the closed-loop dynamics corresponding to this setup is obtained as
\begin{equation} \label{eq:3R-1D2B-p-Dynamics}
    \begin{aligned}
        \begin{bmatrix}
            \dot{p}_{1}
            \\
            \dot{p}_{2}
            \\
            \dot{p}_{3}            
        \end{bmatrix}
        =
        \begin{bmatrix}
            K_{\text{d}} u_{12\text{d}} + K_{\text{d}} u_{13\text{d}}
            \\
            K_{\text{b}} u_{21\text{b}} 
            \\
            K_{\text{b}} u_{31\text{b}} 
        \end{bmatrix}
        ,
    \end{aligned}
\end{equation}
where $ K_{\text{d}} > 0 $ and $ K_{\text{b}} > 0 $ are the gains for the distance and bearing control terms. 
Let $ \left[ \text{L} \right] $ be the dimension for length and $ \left[ \text{T} \right] $ the dimension for time. 
Then we obtain that $ K_{\text{d}} $ has dimension $ \left[ \text{L} \right]^{-2} \left[ \text{T} \right]^{-1} $, and $ K_{\text{b}} $ is expressed in $ \left[ \text{L} \right] \left[ \text{T} \right]^{-1} $.
The \textit{distance control terms} are of the form $ u_{ij\text{d}} = e_{ij\text{d}} z_{ij} $ with the \textit{distance error signal} $ e_{ij\text{d}} = d_{ij}^{2} - \BR{d_{ij}^{\star}}^{2} $.
These are gradient-based control laws obtained from the \textit{distance potential function} $ V_{ij{\text{d}}}\BR{e_{ij\text{d}}} = \frac{1}{4} e_{ij{\text{d}}}^{2} $.
Similarly, the \textit{bearing control terms} are gradient-based control laws obtained from the \textit{bearing potential function} $ V_{ij\text{b}}\BR{e_{ij\text{b}}} = d_{ij} \norm{e_{ij\text{b}}}^{2} $ and are of the form $ u_{ij\text{b}} = e_{ij\text{b}} $ with the \textit{bearing error signal} $ e_{ij\text{b}} = g_{ij} - g_{ij}^{\star} $.
The following findings are reported in \cite{Chan2020} on the stability analysis of the closed-loop system \eqref{eq:3R-1D2B-p-Dynamics}:
\begin{enumerate}
    \item 
    The equilibrium configurations are the desired configurations in which all error signals are zero or zero vector; 
    Furthermore, they are locally asymptotically stable.

    \item 
    The moving configurations occur when the desired distances satisfy $ d_{12}^{\star} \geq \widehat{d} $ and $ d_{13}^{\star} \geq \widehat{d} $ with $ \widehat{d} = \sqrt{3}\sqrt[3]{\frac{R_{\text{bd}}}{2}} $ and gain ratio $ R_{\text{bd}} = \frac{K_{\text{b}}}{K_{\text{d}}} $. 
    The error vector is of the form
    \begin{equation} \label{eq:3R-1D2B-Moving-Conditions}
        \begin{aligned}
            \begin{bmatrix}
                e_{12\text{d}}
                \\
                e_{13\text{d}}
                \\
                e_{12\text{b}}
                \\
                e_{13\text{b}}
            \end{bmatrix}
            =
            -
            \begin{bmatrix}
                \frac{1}{d_{12}} R_{\text{bd}}
                \\
                \frac{1}{d_{13}} R_{\text{bd}}
                \\
                \BR{g_{12}^{\star} + g_{13}^{\star}}
                \\
                \BR{g_{12}^{\star} + g_{13}^{\star}}
            \end{bmatrix}
        \end{aligned}
    \end{equation}
    and the steady-state translation velocity is $ w = K_{\text{b}} \BR{g_{12}^{\star} + g_{13}^{\star}} $.
    It is remarked that these moving configurations have a signed area which is opposite in sign to that of the correct shape.
    When a condition on $ \cos \theta^{\star} $ is satisfied (Lemma 3 in \cite{Chan2020}), we obtained that the linearization matrix has eigenvalues with negative real parts and hence the moving configurations are locally asymptotically stable.  
\end{enumerate}

\subsection{Adding a signed area constraint to \texttt{R1}}
In the works \cite{Anderson2017a, Sugie2018, Cao2019a, Sugie2020}, it was shown that the inclusion of a \textit{signed area constraint} with a proper gain for the resulting control term avoids the occurence of flipped formations. 
Since the moving configurations have a signed area which is opposite to that of the correct formation shape, it can be classified as a flipped formation.  
Hence in the current work, we require \texttt{R1} to additionally fulfill a signed area constraint $ A^{\star} $ involving \texttt{R2} and \texttt{R3}. 
This serves as a strategy to avoid flipped formations and therefore also the moving configurations. 

Recall the expression for the signed area is $ A = \frac{1}{2} z_{12}^{\top} J z_{13} $ with $ J $ being a rotation matrix. 
\texttt{R1} is able to compute this quantity with the available local information. 
For obtaining the control law, we define the \textit{signed area error signal} as $ e_{\text{A}} = A - A^{\star} $. 
The \textit{signed area potential function} is taken as $ V_{\text{A}}\BR{e_{\text{A}}} = e_{\text{A}}^{2} $.
Since \texttt{R1} will be responsible for the signed area constraint, taking the derivative of $ V_{\text{A}} $ with respect to $ p_{1} $ yields $ \PARDIFF{}{p_{1}} V_{\text{A}} = e_{\text{A}} \BR{z_{13} - z_{12}}^{\top} J $.
The gradient-based control law for the signed area task is then $ u_{\text{A}} = e_{\text{A}} J \BR{z_{13} - z_{12}} $.

\subsection{The (\textbf{1D2B}) setup with signed area control term}
Adding the control term $ u_{\text{A}} $ to \texttt{R1} in \eqref{eq:3R-1D2B-p-Dynamics} results in
\begin{equation} \label{eq:3R-1D2B-S-p-Dynamics}
    \begin{aligned}
        \begin{bmatrix}
            \dot{p}_{1}
            \\
            \dot{p}_{2}
            \\
            \dot{p}_{3}            
        \end{bmatrix}
        =
        \begin{bmatrix}
            K_{\text{d}} u_{12\text{d}} + K_{\text{d}} u_{13\text{d}} + K_{\text{A}} u_{\text{A}}
            \\
            K_{\text{b}} u_{21\text{b}} 
            \\
            K_{\text{b}} u_{31\text{b}} 
        \end{bmatrix}
        .
    \end{aligned}
\end{equation}
The control gain for the area task is $ K_{\text{A}} > 0 $ and has dimension $ \left[ \text{L} \right]^{-2} \left[ \text{T} \right]^{-1} $.
Relative to the global coordinate system $ \Sigma^{\text{g}} $, we have $ u_{12\text{b}} = - u_{21\text{b}} $ and $ u_{13\text{b}} = - u_{31\text{b}} $.
The link dynamics corresponding to \eqref{eq:3R-1D2B-S-p-Dynamics} is obtained as 
\begin{equation} \label{eq:3R-1D2B-S-z-Dynamics}
    \begin{aligned}
        \begin{bmatrix}
            \dot{z}_{12}
            \\
            \dot{z}_{13}
            \\
            \dot{z}_{23}
        \end{bmatrix}
        =
        -
        \begin{bmatrix}
            K_{\text{b}} u_{12\text{b}} + K_{\text{d}} u_{12\text{d}} + K_{\text{d}} u_{13\text{d}} + K_{\text{A}} u_{\text{A}}
            \\
            K_{\text{b}} u_{13\text{b}} + K_{\text{d}} u_{12\text{d}} + K_{\text{d}} u_{13\text{d}} + K_{\text{A}} u_{\text{A}}
            \\
            K_{\text{b}} \BR{u_{13\text{b}} - u_{\text{12\text{b}}}}
        \end{bmatrix}
        .
    \end{aligned}
\end{equation}

In the remainder of this paper, we investigate the effect of adding the signed area control term $ u_{\text{A}} $ to the closed-loop system \eqref{eq:3R-1D2B-p-Dynamics}.
In particular,
\begin{enumerate}
    \item 
    we aim to preclude moving configurations by including $ u_{\text{A}} $ and a proper tuning of the associated gain $ K_{\text{A}} $ in relation to other gains; and
    
    \item as 
    including $ u_{\text{A}} $ may introduce other undesired (moving) configurations, we provide conditions which prevent these possible undesired configurations to occur.
\end{enumerate}
We note that the configurations $ \widehat{p} $ yielding the collective error variable $ e = \begin{bmatrix} e_{12\text{d}} & e_{13\text{d}} & e_{\text{A}} & e_{12\text{b}}^{\top} & e_{13\text{b}}^{\top} \end{bmatrix}^{\top} \in \RN^{7} $ to be the zero vector are in the set $ \mathcal{S}_{p} $, i.e., another characterization of the set of desired configurations in terms of the error vector $ e $ is $ \mathcal{S}_{p} = \CBR{p \in \RN^{6} \, | \, e = \mathbb{0}_{7}} $. 

To provide answers to the above determined goals, we are required to solve the following vector equation for the distance pair $ \BR{d_{12}, \, d_{13}} $. 

\begin{proposition} \label{prop:3R-1D2B-S-Condition-EQ-MV-Conf}
    Consider a team of three robots moving according to \eqref{eq:3R-1D2B-S-p-Dynamics}. 
    Define the gain ratios $ R_{\text{bd}} = \frac{K_{\text{b}}}{K_{\text{d}}} > 0 $ and $ R_{\text{Ad}} = \frac{K_{\text{A}}}{K_{\text{d}}} > 0 $.
    Then for the equilibrium and moving configurations, the feasible distance pairs $ \BR{d_{12}, \, d_{13}} $ are solutions to the vector equation
    \begin{equation} \label{eq:3R-1D2B-S-g12-g13-Eq}
        \begin{aligned}
            a g_{12}^{\star} + b g_{13}^{\star} = c g_{12}^{\star \, \perp} + d g_{13}^{\star \, \perp}
            ,
        \end{aligned}
    \end{equation}
    where the coefficients $ a $, $ b $, $ c $, and $ d $ are
    \begin{equation} \label{eq:3R-1D2B-S-EQ-g12-g13-Eq-Coeff}
        \begin{aligned}
            & a = e_{12\text{d}} d_{12}
            , 
            & & b = e_{13\text{d}} d_{13}
            ,
            \\
            & c = - R_{\text{Ad}} e_{\text{A}} d_{12}
            , 
            & & d = R_{\text{Ad}} e_{\text{A}} d_{13}
        \end{aligned}
    \end{equation}
    for equilibrium configurations and 
    \begin{equation} \label{eq:3R-1D2B-S-MV-g12-g13-Eq-Coeff}
        \begin{aligned}
            & a = e_{13\text{d}} d_{13} + R_{\text{bd}}
            , 
            & & b = e_{12\text{d}} d_{12} + R_{\text{bd}}
            ,
            \\
            & c = R_{\text{Ad}} e_{\text{A}} d_{13}
            , 
            & & d = - R_{\text{Ad}} e_{\text{A}} d_{12}
        \end{aligned}
    \end{equation}
    when considering moving configurations.
    
\end{proposition}

\begin{proof}
    First, we consider equilibrium configurations for which $ \dot{p} = \mathbb{0}_{6} $. 
    From the dynamics for the bearing robots \texttt{R2} and \texttt{R3}, we immediately obtain $ e_{12\text{b}} = \mathbb{0}_{2} $ and $ e_{13\text{b}} = \mathbb{0}_{2} $; the bearing constraints are attained. 
    In addition, we have $ d_{12} \neq 0 $ and $ d_{13} \neq 0 $.
    Substituting the correct bearing vectors $ g_{12} = g_{12}^{\star} $ and $ g_{13} = g_{13}^{\star} $ in the dynamics of \texttt{R1} and rearranging the terms result in \eqref{eq:3R-1D2B-S-g12-g13-Eq} with coefficients in \eqref{eq:3R-1D2B-S-EQ-g12-g13-Eq-Coeff}. 
    Next, we consider moving configurations. 
    For this, we focus on the equilibrium points of the link dynamics \eqref{eq:3R-1D2B-S-z-Dynamics} since $ \dot{p}_{i} = \dot{p}_{j} = w \implies \dot{z}_{ij} = \mathbb{0}_{2} $. 
    We have already considered the case $ w = \mathbb{0}_{2} $, so our focus will be on $ w \neq \mathbb{0}_{2} $. 
    Setting $ \dot{z}_{23} = \mathbb{0}_{2} $ leads to $ g_{12} - g_{13} = g_{12}^{\star} - g_{13}^{\star} $.
    The possible solutions are found to be the combinations $ \BR{g_{12}, \, g_{13}} = \BR{g_{12}^{\star}, \, g_{13}^{\star}} $ and $ \BR{g_{12}, \, g_{13}} = \BR{- g_{13}^{\star}, \, - g_{12}^{\star}} $.
    The former corresponds to equilibrium configurations, so $ w = \mathbb{0}_{2} $.
    The latter results in the bearing error signal $ e_{12\text{b}} = e_{13\text{b}} = - \BR{g_{12}^{\star} + g_{13}^{\star}} $.
    Substituting the obtained bearing error signal in \eqref{eq:3R-1D2B-S-z-Dynamics} and rearranging terms, we obtain \eqref{eq:3R-1D2B-S-g12-g13-Eq} with coefficients provided in  \eqref{eq:3R-1D2B-S-MV-g12-g13-Eq-Coeff}.
    Since the bearing vectors are already known, provided $ R_{\text{Ad}} $ and $ R_{\text{bd}} $ are given, the coefficients $ \CBR{a, \, b, \, c, \, d} $ in \eqref{eq:3R-1D2B-S-EQ-g12-g13-Eq-Coeff} and \eqref{eq:3R-1D2B-S-MV-g12-g13-Eq-Coeff} depend solely on the distances $ d_{12} $ and $ d_{13} $.
    This completes the proof.
\end{proof}


Note that pre-multiplying \eqref{eq:3R-1D2B-S-g12-g13-Eq} by a rotation matrix $ Q $ with angle $ \xi $ has no effect on the coefficients $ \CBR{a, \, b, \, c, d} $. 
Hence without loss of generality, we take 
\begin{equation} \label{eq:3R-1D2B-S-g12-g13-Bearing}
    \begin{aligned}
        g_{12}^{\star} = 
        \begin{bmatrix}
            1
            \\
            0
        \end{bmatrix}
        , \:
        g_{13}^{\star} = 
        \begin{bmatrix}
            \cos \theta^{\star} 
            \\
            \sin \theta^{\star}
        \end{bmatrix}
        , \:
        \begin{aligned}
            g_{12}^{\star \, \perp} 
                & = - J g_{12}
                \\
            g_{13}^{\star \, \perp} 
                & = - J g_{13}
        \end{aligned}
        .
    \end{aligned}
\end{equation}
This corresponds to $ \angle g_{12}^{\star} = 0^{\degree} $, $ \angle g_{13}^{\star} = \theta^{\star} $, $ \angle  g_{12}^{\star \, \perp} = 90^{\degree} $, and $ \angle  g_{13}^{\star \, \perp} = 90^{\degree} + \theta^{\star} $, where $ \theta^{\star} $ is the \textit{desired inner angle} enclosed by the bearing vectors $ g_{12}^{\star} $ and $ g_{13}^{\star} $.
When $ \angle g_{12}^{\star} = \alpha \neq 0^{\degree} $, we can pre-multiply \eqref{eq:3R-1D2B-S-g12-g13-Eq} by $ Q\BR{-\alpha} $ to obtain the bearing vectors in \eqref{eq:3R-1D2B-S-g12-g13-Bearing}.
Substituting \eqref{eq:3R-1D2B-S-g12-g13-Bearing} in \eqref{eq:3R-1D2B-S-g12-g13-Eq} yields the set of equations
\begin{equation} \label{eq:3R-1D2B-S-g12-g13-Eq-Set}
    \left \{
    \begin{aligned}
        a + b \cos \theta^{\star} 
            & = - d \sin \theta^{\star}
        \\
        b \sin \theta^{\star} 
            & = c + d \cos \theta^{\star}
            .
    \end{aligned}
    \right.
\end{equation}
Equivalently, we obtain 
\begin{equation} \label{eq:3R-1D2B-S-g12-g13-Eq-Set-V2}
    \left \{
    \begin{aligned}
        a \sin \theta^{\star}
            & = - c \cos \theta^{\star} - d
        \\
        b \sin \theta^{\star} 
            & = c + d \cos \theta^{\star}
            .
    \end{aligned}
    \right.    
\end{equation}
In the forthcoming analysis, we aim to find feasible distance pairs $ \BR{d_{12}, \, d_{13}} $ in Proposition \ref{prop:3R-1D2B-S-Condition-EQ-MV-Conf} by solving \eqref{eq:3R-1D2B-S-g12-g13-Eq-Set-V2}.
We assume $ \theta^{\star} $ is in the region $ \theta^{\star} \in \BR{0^{\degree}, 180^{\degree}} $; in this case, the robots are ordered in a counter-clockwise setting and hence the \textit{desired} signed area $ A^{\star} $ is positive.

\section{Analysis on Isosceles Triangles} \label{sec:Isosceles-Triangle-Analysis}
As a first endeavor, we focus on solving the set of equations \eqref{eq:3R-1D2B-S-g12-g13-Eq-Set-V2} for the class of \textit{isosceles} triangles. 
An isosceles triangle has two equal sides of length  $ \ell > 0 $ and two equal angles with value $ \gamma \in \BR{0^{\degree}, \, 90^{\degree}} $. 
The equal sides are called legs and the third side is the base. 
The equal angles are called base angles and the angle included by the legs is the vertex angle. 

In the current setup, we assume the distance constraints $ d_{12}^{\star} $ and $ d_{13}^{\star} $ are equal, i.e., $ d_{12}^{\star} = d_{13}^{\star} = \ell $ are the legs of the triangle and the desired inner angle $ \theta^{\star} $ is the vertex angle and satisfies $ \theta^{\star} = 180^{\degree} - 2\gamma $. 
The isosceles triangle is \textit{acute} when $ \theta^{\star} \in \BR{0^{\degree}, \, 90^{\degree}} $, \textit{right} when $ \theta^{\star} = 90^{\degree} $, and \textit{obtuse} when $ \theta^{\star} \in \BR{90^{\degree}, \, 180^{\degree}} $.

We first obtain the set of equations \eqref{eq:3R-1D2B-S-g12-g13-Eq-Set-V2} for the equilibrium and moving configurations corresponding to this class of triangles. 
To this end, we
parametrize the actual distances $d_{12}$ and $d_{13}$ by 
\begin{equation} \label{eq:3R-1D2B-S-ISO-EQ-PAR-XY}
    d_{12} = x \ell, \, d_{13} = y \ell \quad \text{ with } x, \, y > 0.
\end{equation}
The value $ x = 1 $ means robot \texttt{R1} satisfies the distance constraint relative to \texttt{R2}. 
A similar result concerning \texttt{R3} holds when $ y = 1 $.

\subsection{Equilibrium configurations}
For equilibrium configurations, we recall that according to Proposition \ref{prop:3R-1D2B-S-Condition-EQ-MV-Conf}, the bearing robots \texttt{R2} and \texttt{R3} attain its individual bearing task, i.e. $ g_{12} = g_{12}^{\star} $ and $ g_{13} = g_{13}^{\star} $. 
It follows then that $ \sin \theta = \sin \theta^{\star} $.
With the parametrization in \eqref{eq:3R-1D2B-S-ISO-EQ-PAR-XY}, the distance and signed area error signals evaluate to $ e_{12\text{d}} = \BR{x^{2} - 1} \ell^{2} $, $ e_{13\text{d}} = \BR{y^{2} - 1} \ell^{2} $, and $ e_{\text{A}} = \frac{1}{2} \sin \theta^{\star} \BR{xy - 1} \ell^{2} $.
Substituting these relations into \eqref{eq:3R-1D2B-S-g12-g13-Eq-Set-V2} with coefficients $ \CBR{a, \, b, \, c, \, d} $ defined in \eqref{eq:3R-1D2B-S-EQ-g12-g13-Eq-Coeff} yields 
\begin{equation} \label{eq:3R-1D2B-S-ISO-EQ-g12-g13-Set}
    \left \{
    \begin{aligned}
        \BR{x^{2} - 1} x
            & = \frac{1}{2} R_{\text{Ad}} \BR{xy - 1} \BR{x \cos \theta^{\star} - y}
        \\
        \BR{y^{2} - 1} y
            & = \frac{1}{2} R_{\text{Ad}} \BR{xy - 1} \BR{y \cos \theta^{\star} - x}
            .
    \end{aligned}
    \right.    
\end{equation}
We find out whether solutions of the form $ x = y $ and $ x \neq 1 $ are feasible.
In this case, expression \eqref{eq:3R-1D2B-S-ISO-EQ-g12-g13-Set} reduces to $ \BR{1 + \frac{1}{2} R_{\text{Ad}} \BR{1 - \cos \theta^{\star}}} x = 0 $.
For it to hold, we require $ R_{\text{Ad}} = - \frac{2}{1 - \cos \theta^{\star}} < -1 $ or $ x = 0 $.
Since $ R_{\text{Ad}} > 0 $ and $ x > 0 $, it follows both conditions cannot be met.
By De Morgan's laws, solutions to \eqref{eq:3R-1D2B-S-ISO-EQ-g12-g13-Set} are either of the form $ x \neq y $ or $ x = 1 $.
When $ x \neq y $, subtracting the equations in \eqref{eq:3R-1D2B-S-ISO-EQ-g12-g13-Set} results in 
\begin{equation} \label{eq:3R-1D2B-S-ISO-EQ-g12-g13-Diff}
    \begin{aligned}
        x^{2} + y^{2} + xy - 1
            & = \frac{1}{2} R_{\text{Ad}} \BR{1 + \cos \theta^{\star}} \BR{xy - 1}
            .
    \end{aligned}
\end{equation}
In \eqref{eq:3R-1D2B-S-ISO-EQ-g12-g13-Set}, we see the presence of $ \cos \theta^{\star} $ terms.
For acute isosceles triangles, $ \cos \theta^{\star} \in \BR{0, \, 1} $ while for right and obtuse isosceles triangles, $ \cos \theta^{\star} \in \left(-1, \, 0\right] $. 
In the forthcoming sections, we will divide the analysis in these two sub-regions. 
We state the following result which holds for both sub-regions of $ \cos \theta^{\star} $:

\begin{proposition} \label{prop:3R-1D2B-S-ISO-EQ-Conditions-I}
    For robot \texttt{R1}, satisfying one of its assigned tasks is equivalent to satisfying all its assigned tasks. 
    In particular, with the parametrization of the distances in \eqref{eq:3R-1D2B-S-ISO-EQ-PAR-XY}, we have 
    \begin{enumerate}
        \item 
        $ x = 1 \iff y = 1 \, \wedge \, xy = 1 $;
        
        \item 
        $ y = 1 \iff x = 1 \, \wedge \, xy = 1 $;

        \item 
        $ xy = 1 \iff x = 1 \, \wedge \, y = 1 $.
    \end{enumerate}
\end{proposition}

\begin{proof}
    The necessity part $ \BR{\impliedby} $ is immediately observed for all the three statements, so we focus only on the sufficient part $ \BR{\implies} $.
    \begin{enumerate}
        \item 
        $ x = 1 \implies y = 1 \, \wedge \, xy = 1 $;
        ~\\
        Substituting $ x = 1 $ in \eqref{eq:3R-1D2B-S-ISO-EQ-g12-g13-Set} yields 
        \begin{equation} \label{eq:3R-1D2B-S-ISO-EQ-g12-g13-Set-Case-1}
            \left \{
            \begin{aligned}
                0
                    & = \frac{1}{2} R_{\text{Ad}} \BR{y - 1} \BR{\cos \theta^{\star} - y}
                    \\
                \BR{y^{2} - 1} y
                    & = \frac{1}{2} R_{\text{Ad}} \BR{y - 1} \BR{y\cos \theta^{\star} - 1}
                    .
            \end{aligned}
            \right.    
        \end{equation}
        The first equation in \eqref{eq:3R-1D2B-S-ISO-EQ-g12-g13-Set-Case-1} is satisfied when $ y = 1 $ or $ y = \cos \theta^{\star} $. 
        The option $ y = 1 $ holds for the second equation. 
        In addition, $ xy = 1 $. 
        Since we know $ y > 0 $, it follows that option $ y = \cos \theta^{\star} $ is feasible only when $ \cos \theta^{\star} > 0 $.
        Substituting $ y = \cos \theta^{\star} $ in the second equation yields $ \cos \theta^{\star} = - \frac{1}{2} R_{\text{Ad}} \BR{1 - \cos \theta^{\star}} $. 
        The left-hand side (LHS) is positive while the right-hand side (RHS) is negative since $ R_{\text{Ad}} > 0 $. 
        We infer that $ y = \cos \theta^{\star} $ does not satisfy the second equation and hence it is not a solution to \eqref{eq:3R-1D2B-S-ISO-EQ-g12-g13-Set-Case-1}.
        
        \item 
        $ y = 1 \implies x = 1 \, \wedge \, xy = 1 $;
        ~\\
        Substituting $ y = 1 $ in \eqref{eq:3R-1D2B-S-ISO-EQ-g12-g13-Set} yields 
        \begin{equation} \label{eq:3R-1D2B-S-ISO-EQ-g12-g13-Set-Case-2}
            \left \{
            \begin{aligned}
                \BR{x^{2} - 1} x
                    & = \frac{1}{2} R_{\text{Ad}} \BR{x - 1} \BR{x \cos \theta^{\star} - 1}
                    \\
                0
                    & = \frac{1}{2} R_{\text{Ad}} \BR{x - 1} \BR{\cos \theta^{\star} - x}
                    .
            \end{aligned}
            \right.    
        \end{equation}
        The second equation in \eqref{eq:3R-1D2B-S-ISO-EQ-g12-g13-Set-Case-2} is satisfied when $ x = 1 $ or $ x = \cos \theta^{\star}  $. 
        The option $ x = 1 $ holds for the first equation. 
        In addition, $ xy = 1 $. 
        Since we know $ x > 1 $, it follows that option $ x = \cos \theta^{\star} $ is feasible only when $ \cos \theta^{\star} > 0 $.
        Substituting $ x = \cos \theta^{\star} $ in the first equation yields $ \cos \theta^{\star} = - \frac{1}{2} R_{\text{Ad}} \BR{1 - \cos \theta^{\star}} $. 
        The LHS is positive while the RHS is negative since $ R_{\text{Ad}} > 0 $. 
        We infer that $ x = \cos \theta^{\star} $ does not satisfy the first equation and hence it is not a solution to \eqref{eq:3R-1D2B-S-ISO-EQ-g12-g13-Set-Case-2}.

        \item 
        $ xy = 1 \implies x = 1 \, \wedge \, y = 1 $;
        ~\\
        Substituting $ xy = 1 $ in \eqref{eq:3R-1D2B-S-ISO-EQ-g12-g13-Set} yields 
        \begin{equation} \label{eq:3R-1D2B-S-ISO-EQ-g12-g13-Set-Case-3}
            \left \{
            \begin{aligned}
                \BR{x^{2} - 1} x
                    & = 0
                    \\
                \BR{y^{2} - 1} y
                    & = 0
            \end{aligned}
            \right.   
            \iff
            \left \{
            \begin{aligned}
                x
                    & = \pm 1
                    \\
                y
                    & = \pm 1
                    .
            \end{aligned}
            \right.   
        \end{equation} 
        Since $ x > 0 $ and $ y > 0 $ holds, the only possible combination is $ x = 1 \, \wedge \, y = 1 $.
    \end{enumerate}
    This completes the proof. 
\end{proof}

In Proposition \ref{prop:3R-1D2B-S-ISO-EQ-Conditions-I}, at least one of the tasks assigned to \texttt{R1} is attained. 
It remains to investigate the case when none of the assigned tasks is achieved, i.e., the case $ x \neq 1 $, $ y \neq 1 $, $ xy \neq 1 $, and $ x \neq y $ by De Morgan's laws for the sub-regions of $ \cos \theta^{\star} $. 
We will deal with this in the forthcoming sections. 

\subsection{Moving configurations}
Previously, in Proposition \ref{prop:3R-1D2B-S-Condition-EQ-MV-Conf}, we have obtained that the bearing vectors corresponding to moving formations are $ g_{12} = - g_{13}^{\star} $ and $ g_{13} = - g_{12}^{\star} $. 
It follows that $ \sin \theta = - \sin \theta^{\star} $; the formation is flipped and rotated. 
With the parametrization of the distances $ d_{12} = x \ell $ and $ d_{13} = y \ell $ in \eqref{eq:3R-1D2B-S-ISO-EQ-PAR-XY}, we obtain that distance error signals are the same as before while the signed area error signal evaluates to $ e_{\text{A}} = \frac{1}{2} \sin \theta^{\star} \BR{-xy - 1} \ell^{2} $ for moving configurations.
The set of equations \eqref{eq:3R-1D2B-S-g12-g13-Eq-Set-V2} with coefficients $ \CBR{a, \, b, \, c, \, d} $ in \eqref{eq:3R-1D2B-S-MV-g12-g13-Eq-Coeff} is found to be
\begin{equation} \label{eq:3R-1D2B-S-ISO-MV-g12-g13-Set} 
    \left \{
    \begin{aligned}
        \BR{x^{2} - 1} x \ell^{3} + R_{\text{bd}}
            & = \frac{1}{2} R_{\text{Ad}} \BR{xy + 1} \BR{x \cos \theta^{\star} - y} \ell^{3}
        \\
        \BR{y^{2} - 1} y \ell^{3} + R_{\text{bd}}
            & = \frac{1}{2} R_{\text{Ad}} \BR{xy + 1} \BR{y \cos \theta^{\star} - x} \ell^{3}
        .
    \end{aligned}
    \right.    
\end{equation}
Again, we find out whether solutions of the form $ x = y $ are feasible. 
With $ x = y $, \eqref{eq:3R-1D2B-S-ISO-MV-g12-g13-Set} reduces to 
\begin{equation} \label{eq:3R-1D2B-S-ISO-MV-g12-g13-Equal}
    \begin{aligned}
        & 
        \BR{x^{2} - 1} x \ell^{3} + R_{\text{bd}}
            = -\frac{1}{2} R_{\text{Ad}} \BR{1 - \cos \theta^{\star}} \BR{x^{2} + 1} x \ell^{3}
        .
    \end{aligned}
\end{equation}
Observe that the RHS of \eqref{eq:3R-1D2B-S-ISO-MV-g12-g13-Equal} is negative; for the LHS to be negative, $ x < 1 $ is required. 
The exact range for $ x $ is provided in Corollary \ref{cor:3R-1D2B-S-MV-Cubic-Equation-r}.
The difference equation for \eqref{eq:3R-1D2B-S-ISO-MV-g12-g13-Set} is with $ x \neq y $,
\begin{equation} \label{eq:3R-1D2B-S-ISO-MV-g12-g13-Diff}
    \begin{aligned}
        & 
        x^{2} + y^{2} + xy - 1
            = \frac{1}{2} R_{\text{Ad}} \BR{1 + \cos \theta^{\star}} \BR{xy + 1}
        .
    \end{aligned}
\end{equation}
Similar to the equilibrium configurations, we will divide the forthcoming analysis on moving configurations in two sub-regions, namely acute isosceles triangles with $ \cos \theta^{\star} > 0 $ and right and obtuse isosceles triangles having $ \cos \theta^{\star} \leq 0 $.
Before getting into these analyses, we state the following result for the LHS of \eqref{eq:3R-1D2B-S-ISO-MV-g12-g13-Set}:

\begin{proposition} \label{prop:3R-1D2B-S-ISO-MV-Cubic-Equation-z}
    Given a cubic equation of the form 
    \begin{equation} \label{eq:3R-1D2B-S-ISO-MV-Case-I-2-Cubic}
        \begin{aligned}
            f\BR{\mathbb{z}}
                \coloneqq \mathbb{z}^{3} - \ell^{2} \mathbb{z} + R_{\text{bd}}
            ,
        \end{aligned}
    \end{equation}
    where $ \mathbb{z} = \mathbb{r} \ell $ denotes a general variable for the distance and $ \mathbb{r} \in \CBR{x, \, y} $.
    Let $ \widehat{d} = \sqrt{3} \sqrt[3]{\frac{R_{\text{bd}}}{2}} $.
    Then for $ \mathbb{z} > 0 $, the cubic equation in \eqref{eq:3R-1D2B-S-ISO-MV-Case-I-2-Cubic} takes values
    \begin{equation} \label{eq:3R-1D2B-S-ISO-MV-Case-I-2-Cubic-Value}
        f\BR{\mathbb{z}} = 
        \begin{cases}
            > 0 & \text{if } \ell < \widehat{d}
            \\
            \geq 0 & \text{if } \ell = \widehat{d}
            \\
            \gtreqqless 0 & \text{if } \ell > \widehat{d}
            .
        \end{cases}
    \end{equation}
\end{proposition}

\begin{proof}
    By comparison with \eqref{eq:Prel-Reduced-Cubic}, we obtain the coefficients $ c = - \ell^{2} $ and $ d = R_{\text{bd}} $ for \eqref{eq:3R-1D2B-S-ISO-MV-Case-I-2-Cubic}. 
    The discriminant evaluates to $ \Delta = 4 \ell^{6} - 27 R_{\text{bd}}^{2} $.
    Since $ c < 0 $ and $ d > 0 $, applying Lemma \ref{lem:Prel-Reduced-Cubic-Roots-Positive} yields for $ \Delta \geq 0 \iff \ell \geq \widehat{d} $ the positive roots 
    \begin{equation} \label{eq:3R-1D2B-S-ISO-MV-Case-I-2-Cubic-Pos-Roots}
        \begin{aligned}
            \mathbb{z}_{\text{p}1}
                & = \frac{2}{3} \sqrt{3} \cos \BR{\frac{1}{3} \varphi - 120^{\degree}} \ell 
                & \in \left(0, \, \frac{1}{3}\sqrt{3} \right] \ell 
                \\
            \mathbb{z}_{\text{p}2}
                & = \frac{2}{3} \sqrt{3} \cos \BR{\frac{1}{3} \varphi} \ell 
                & \in  \left[\frac{1}{3}\sqrt{3}, \, 1\right) \ell 
                ,
        \end{aligned}
    \end{equation}
    where $ \varphi = \tan^{-1} \BR{-2 R_{\text{bd}}^{-1} \sqrt{\frac{\Delta}{108}}} \in \left(90^{\degree}, \, 180^{\degree} \right] $.
    Notice that $ \varphi $ depends on both the desired length $ \ell $ and the gain ratio $ R_{\text{bd}} $.
    Before considering the different sub-regions for $ \ell $, we also compute the derivative of \eqref{eq:3R-1D2B-S-ISO-MV-Case-I-2-Cubic}, yielding $ f'\BR{\mathbb{z}} = 3 \mathbb{z}^{2} - \ell^{2} $.
    The roots are $ f'\BR{\mathbb{z}} = 0 \iff \mathbb{z} = \pm \frac{1}{3} \sqrt{3} \ell $.
    From the first derivative test, the maximum and minimum are found to be $ f_{\max} = f\BR{- \frac{1}{3}\sqrt{3} \ell} = \frac{2}{9} \sqrt{3} \ell^{3} + R_{\text{bd}} > 0 $ and $ f_{\min} = f\BR{\frac{1}{3}\sqrt{3} \ell} = - \frac{2}{9} \sqrt{3} \ell^{3} + R_{\text{bd}} \gtreqqless 0 $.
    The sign of $ f_{\min} $ depends on the value for $ \ell $.
    In addition, $ f\BR{0} = f \BR{\ell} = R_{\text{bd}} > 0 $.
    Now we are ready to consider the different sub-regions of $ \ell $ for $ \mathbb{z} > 0 $:
    \begin{enumerate}
        \item 
        $ \ell < \widehat{d} $;
        ~\\
        We only have one local minimum for $ f\BR{\mathbb{z}} $ in the positive range. 
        With $ \ell < \widehat{d} $, it follows $ \ell^{3} < \frac{3}{2} \sqrt{3} R_{\text{bd}} $. 
        Correspondingly, we have $ f_{\min} = - \frac{2}{9} \sqrt{3} \ell^{3} + R_{\text{bd}} > 0 $ implying that $ f\BR{\mathbb{z}} > 0 $ for all $ \mathbb{z} > 0 $.
            
        \item 
        $ \ell = \widehat{d} $;
        ~\\
        With $ \Delta = 0 $, we obtain $ \varphi = 180^{\degree} $; the positive roots \eqref{eq:3R-1D2B-S-ISO-MV-Case-I-2-Cubic-Pos-Roots} are equal and have value $ \mathbb{z}_{\text{p}1} = \mathbb{z}_{\text{p}2} = \frac{1}{3} \sqrt{3} \ell $.
        Also, the minimum of $ f\BR{\mathbb{z}} $ occurs at this point, i.e., $ f\BR{\mathbb{z}_{\text{p1}}} = f\BR{\mathbb{z}_{\text{p2}}} = 0 = f_{\min} $.
        We thus have $ f\BR{\mathbb{z}} \geq 0 $ for all $ \mathbb{z} > 0$. 
        
        \item 
        $ \ell > \widehat{d} $;
        ~\\
        For $ \ell > \widehat{d} $, we have two distinct positive roots in \eqref{eq:3R-1D2B-S-ISO-MV-Case-I-2-Cubic-Pos-Roots}.
        Also, it follows $ \ell^{3} > \frac{3}{2} \sqrt{3} R_{\text{bd}} $. 
        Correspondingly, we have $ f_{\min} = - \frac{2}{9} \sqrt{3} \ell^{3} + R_{\text{bd}} < 0 $.
        We note that $ \mathbb{z} = \frac{1}{3}\sqrt{3} $ lies in the region $ \BR{\mathbb{z}_{\text{p}1}, \, \mathbb{z}_{\text{p}2}} $.
        The function $ f\BR{\mathbb{z}} $ thus takes values 
        \begin{equation} \label{eq:3R-1D2B-S-ISO-MV-Case-I-2-Cubic-Value-l>d}
            f\BR{\mathbb{z}} 
            = 
            \begin{cases}
                > 0 & \text{if } \mathbb{z} \in \BR{0, \, \mathbb{z}_{\text{p}1}} \cup \BR{\mathbb{z}_{\text{p}2}, \, \infty}
                \\
                = 0 & \text{if } \mathbb{z} \in \CBR{\mathbb{z}_{\text{p}1}, \, \mathbb{z}_{\text{p}2}}
                \\
                < 0 & \text{if } \mathbb{z} \in \BR{\mathbb{z}_{\text{p}1}, \, \mathbb{z}_{\text{p}2}}
                .
            \end{cases}
        \end{equation}
    \end{enumerate}
    This completes the proof. 
\end{proof}

Since $ \mathbb{z} = \mathbb{r} \ell $, we obtain the following corollary:

\begin{corollary} \label{cor:3R-1D2B-S-MV-Cubic-Equation-r}
    The cubic function $ f $ in \eqref{eq:3R-1D2B-S-ISO-MV-Case-I-2-Cubic} takes on values $ f\BR{\mathbb{r}} > 0 $ when $ \ell < \widehat{d} $, $ f\BR{\mathbb{r}} \geq 0 $ when $ \ell = \widehat{d} $ and for $ \ell > \widehat{d} $, we have 
    \begin{equation} \label{eq:3R-1D2B-S-ISO-MV-Case-I-2-Cubic-Value-l>d-r}
        f\BR{\mathbb{r}} 
        = 
        \begin{cases}
            > 0 & \text{if } \mathbb{r} \in \BR{0, \, \mathbb{r}_{1}} \cup \BR{\mathbb{r}_{2}, \, \infty}
            \\
            = 0 & \text{if } \mathbb{r} \in \CBR{\mathbb{r}_{1}, \, \mathbb{r}_{2}}
            \\
            < 0 & \text{if } \mathbb{r} \in \BR{\mathbb{r}_{1}, \, \mathbb{r}_{2}}
            ,
        \end{cases}
    \end{equation}
    where $ \mathbb{r}_{1} = \frac{2}{3} \sqrt{3} \cos \BR{\frac{1}{3} \varphi - 120^{\degree}} \in \left(0, \, \frac{1}{3}\sqrt{3} \right) $, $ \mathbb{r}_{2} = \frac{2}{3} \sqrt{3} \cos \BR{\frac{1}{3} \varphi} \in \left(\frac{1}{3}\sqrt{3}, \, 1 \right) $, and
    \\ $ \varphi\BR{\ell, \, R_{\text{bd}}} = \tan^{-1} \BR{-2 R_{\text{bd}}^{-1} \sqrt{\frac{\Delta}{108}}} \in \left(90^{\degree}, \, 180^{\degree} \right) $.
\end{corollary}

In the upcoming sections, we will study more in detail the set of equations for the equilibrium and the moving configurations for acute, right, and obtuse isosceles triangles. 

\section{Acute Isosceles Triangles} \label{sec:Isosceles-Triangle-Analysis-Acute}
Herein, we focus on acute isosceles triangles in which the vertex angle $ \theta^{\star} $ is in the range $ \theta^{\star} \in \BR{0, \, 90^{\degree}} \iff \cos \theta^{\star} \in \BR{0, \, 1} $.

\subsection{Equilibrium configurations}
Previously, we have shown in Proposition \ref{prop:3R-1D2B-S-ISO-EQ-Conditions-I} that when distance robot \texttt{R1} attains one of its assigned tasks, it is equivalent to attaining all its assigned tasks. 
We investigate now whether there exist equilibrium configurations in which none of the tasks assigned to \texttt{R1} is attained. 
The following proposition provides necessary conditions which the variables $ x $, $ y $, and the product $ xy $ are required to satisfy:

\begin{proposition} \label{prop:3R-1D2B-S-ISO-EQ-Conditions-II}
    Assume $ x \neq 1 $, $ y \neq 1 $, $ xy \neq 1 $, and $ x \neq y $, where $ x $ and $ y $ are defined in \eqref{eq:3R-1D2B-S-ISO-EQ-PAR-XY}.
    Define $ \mathbb{A} = x \cos \theta^{\star} - y $ and $ \mathbb{B} = y \cos \theta^{\star} - x $.
    For the existence of equilibrium configurations in \eqref{eq:3R-1D2B-S-ISO-EQ-g12-g13-Set} satisfying the given constraints, we require that $ xy - 1 > 0 $ in combination with either $ 1). $ $ x > 1 $, $ y < 1 $, and $ \mathbb{A} > 0 $ or $ 2). $ $ x < 1 $, $ y > 1 $, and $ \mathbb{B} > 0 $.
\end{proposition}

\begin{proof}
    By direct computation, the following two relations hold: $ 1). $ $ \mathbb{A} \geq 0 \implies \mathbb{B} < 0 $, and $ 2). $ $ \mathbb{B} \geq 0 \implies \mathbb{A} < 0 $. 
    Due to $ x \neq 1 $, we cannot have $ \mathbb{A} = 0 $. 
    Similarly, with $ y \neq 1 $, $ \mathbb{B} = 0 $ does not hold. 
    Also, the combination $ \mathbb{A} < 0 $ and $ \mathbb{B} < 0 $ cannot hold since assuming $ xy - 1 > 0 $ on the RHS of \eqref{eq:3R-1D2B-S-ISO-EQ-g12-g13-Set} yields $ x < 1 $ and $ y < 1 $ on the LHS. 
    With $ x < 1 $ and $ y < 1 $, we obtain $ xy < 1 $ and this contradicts the assumption $ xy - 1 > 0 $. 
    Similar argument holds when $ xy - 1 < 0 $ is taken. 
    The remaining feasible combinations are then $ \mathbb{A} > 0 \, \wedge \, \mathbb{B} < 0 $ and $ \mathbb{A} < 0 \, \wedge \, \mathbb{B} > 0 $.
    It follows that on the LHS, we either have the combination $ x < 1 \, \wedge \, y > 1 $ or $ x > 1 \, \wedge \, y < 1 $.
    This depends on the sign of $ xy - 1 $ as follows:
    \begin{enumerate}
        \item 
        $ xy - 1 > 0 $;
            \begin{enumerate}
                \item $ \mathbb{A} > 0 \, \wedge \, \mathbb{B} < 0 $ results in $ x > 1 \, \wedge \, y < 1 $.
                
                \item $ \mathbb{A} < 0 \, \wedge \, \mathbb{B} > 0 $ results in $ x < 1 \, \wedge \, y > 1 $.
            \end{enumerate}
        
        \item 
        $ xy - 1 < 0 $;
            \begin{enumerate}
                \item $ \mathbb{A} > 0 \, \wedge \, \mathbb{B} < 0 $ results in $ x < 1 \, \wedge \, y > 1 $.
                
                \item $ \mathbb{A} < 0 \, \wedge \, \mathbb{B} > 0 $ results in $ x > 1 \, \wedge \, y < 1 $.
            \end{enumerate}
    \end{enumerate}
    Consider now the cases where $ xy - 1 < 0 $. 
    Assuming $ \mathbb{A} > 0 $ leads to $ y < x $. 
    However, we have found $ x < 1 \, \wedge \, y > 1 $ on the LHS implying $ x < y $ and hence a contradiction. 
    With $ \mathbb{B} > 0 $, we infer $ x < y $ while based on the signs of $ x $ and $ y $ on the LHS, we have $ y < x $ and again a contradiction. 
    For $ xy - 1 > 0 $, we can find feasible values for $ x $ and $ y $ satisfying the listed constraints. 
    This completes the proof.
\end{proof}

A proper choice for the gain ratio $ R_{\text{Ad}} $ can prevent the occurrence of equilibrium points satisfying the conditions in Proposition \ref{prop:3R-1D2B-S-ISO-EQ-Conditions-II}:

\begin{lemma} \label{lem:3R-1D2B-S-ISO-EQ-Main-Acute}
    Consider a team of three robots moving according to \eqref{eq:3R-1D2B-S-p-Dynamics} with $ K_{\text{d}} > 0 $, $ K_{\text{b}} > 0 $, and $ K_{\text{A}} > 0 $.
    Define the gain ratio $ R_{\text{Ad}} = \frac{K_{\text{A}}}{K_{\text{d}}} $.
    Furthermore, let the desired formation shape be an isosceles triangle with legs $ \ell > 0 $ and vertex angle $ \theta^{\star} \in \BR{0^{\degree}, \, 90^{\degree}} $.
    Finally, parametrize the distances $ d_{12} $ and $ d_{13} $ as in \eqref{eq:3R-1D2B-S-ISO-EQ-PAR-XY}.
    If $ R_{\text{Ad}} \leq \frac{6}{1 + \cos \theta^{\star}} $, then the equilibrium configurations $ p_{\text{eq}} $ are the desired ones in $ \mathcal{S}_{p} $ in which all the individual assigned tasks are attained.
\end{lemma}

\begin{proof}
    We first consider \eqref{eq:3R-1D2B-S-ISO-EQ-g12-g13-Diff}.
    Rearranging the terms yields 
    \begin{equation} \label{eq:3R-1D2B-S-ISO-EQ-g12-g13-Diff-V2}
        \begin{aligned}
            x^{2} + y^{2} + \BR{1 - \mathbb{d}} \BR{xy - 1}
                & = 0
                ,
        \end{aligned}
    \end{equation}
    where $ \mathbb{d} = \frac{1}{2} R_{\text{Ad}} \BR{1 + \cos \theta^{\star}} > 0 $. 
    Since $ x > 0 $, $ y > 0 $, and $ xy - 1 > 0 $ from Proposition \ref{prop:3R-1D2B-S-ISO-EQ-Conditions-II}, it follows that $ 1 - \mathbb{d} \geq 0 \iff \mathbb{d} \leq 1 $ will yield the LHS to be positive and hence no feasible combination $ \BR{x, \, y} $ for \eqref{eq:3R-1D2B-S-ISO-EQ-g12-g13-Diff-V2}. 
    Adding and subtracting $ 2 xy $ to \eqref{eq:3R-1D2B-S-ISO-EQ-g12-g13-Diff-V2} yields 
    \begin{equation} \label{eq:3R-1D2B-S-ISO-EQ-g12-g13-Diff-V3}
        \BR{x - y}^{2} + \BR{3 - \mathbb{d}} xy - \BR{1 - \mathbb{d}} = 0.
    \end{equation}
    Choosing $ \mathbb{d} $ in the range $ \left(1, \, 3\right] $, we obtain that all terms on the LHS are non-negative and at least one term is positive; their sum is then also positive and hence we have no feasible combination $ \BR{x, \, y} $ for \eqref{eq:3R-1D2B-S-ISO-EQ-g12-g13-Diff-V3}. 
    Notice that $ \mathbb{d} \leq 3 \iff R_{\text{Ad}} \leq \frac{6}{1 + \cos \theta^{\star}} $.
    Since solutions to \eqref{eq:3R-1D2B-S-ISO-EQ-g12-g13-Set} should naturally satisfy \eqref{eq:3R-1D2B-S-ISO-EQ-g12-g13-Diff}, we infer that provided $ R_{\text{Ad}} \leq \frac{6}{1 + \cos \theta^{\star}} $, we do not have a feasible combination $ \BR{x, \, y} $ with $ x \neq y $ satisfying \eqref{eq:3R-1D2B-S-ISO-EQ-g12-g13-Set}.
    The only possible combination for $ x = y $ is the pair $ \BR{x, \, y} = \BR{1, \, 1} $ which corresponds to robot \texttt{R1} satisfying all its assigned tasks.
    As robots \texttt{R2} and \texttt{R3} also attain its individual task, we conclude that all robots in the team attain its individual tasks; i.e., $ e = \mathbb{0}_{7} $.
    This completes the proof. 
\end{proof}

Since we are dealing with acute isosceles triangles, we obtain that the upper bound for $ R_{\text{Ad}} $ in Lemma \ref{lem:3R-1D2B-S-ISO-EQ-Main-Acute} is in the range $ \frac{6}{1 + \cos \theta^{\star}} \in \BR{3, \, 6} $.
This seems rather limited. 
Moreover, in obtaining feasible ranges for $ R_{\text{Ad}} $ in Lemma \ref{lem:3R-1D2B-S-ISO-EQ-Main-Acute}, we only made use of the condition that the product $ xy $ is larger than $ 1 $ while no specific constraints on $ x $ and $ y $ in Proposition \ref{prop:3R-1D2B-S-ISO-EQ-Conditions-II} were utilized. 
We could ask ourselves $ 1). $ whether we can expand the region of $ R_{\text{Ad}} $ by taking into account all the conditions provided in Proposition \ref{prop:3R-1D2B-S-ISO-EQ-Conditions-II}. 
If there are still feasible combinations $ \BR{x, \, y} $ solving \eqref{eq:3R-1D2B-S-ISO-EQ-g12-g13-Diff} while at the same time satisfying all the conditions in Proposition \ref{prop:3R-1D2B-S-ISO-EQ-Conditions-II} with $ \mathbb{d} > 3 $, then a follow-up question would be $ 2). $ whether these particular combinations $ \BR{x, \, y} $ would solve the set of equations \eqref{eq:3R-1D2B-S-ISO-EQ-g12-g13-Set}. 

For now, we focus on the set of constraints $ x > 1 $, $ y < 1 $, $ \mathbb{A} > 0 $, and $ xy -1 > 0 $ in Proposition \ref{prop:3R-1D2B-S-ISO-EQ-Conditions-II}.
For a fixed value $ x = \bar{x} > 1 $, we obtain as solutions to \eqref{eq:3R-1D2B-S-ISO-EQ-g12-g13-Diff-V2}
\begin{equation} \label{eq:3R-1D2B-S-ISO-EQ-g12-g13-Diff-Roots}
    y = \frac{\mathbb{d} - 1}{2} \bar{x} \pm \frac{1}{2} \sqrt{\BR{\BR{\mathbb{d} + 1} \BR{\mathbb{d} - 3}}\bar{x}^{2} - 4\BR{\mathbb{d} - 1}}
    .
\end{equation}
With $ \mathbb{d} > 3 $, we have $ \frac{\mathbb{d} - 1}{2} > 1 $. 
The feasible value for $ y $ which could satisfy $ y < 1 $ is then $ \bar{y} = \mathbb{a} \bar{x} - \mathbb{b} $ with $ \mathbb{a} = \frac{\mathbb{d} - 1}{2} $ and $ \mathbb{b} = \frac{1}{2} \sqrt{\BR{\BR{\mathbb{d} + 1} \BR{\mathbb{d} - 3}}\bar{x}^{2} - 4\BR{\mathbb{d} - 1}} $ since the alternative $ y = \mathbb{a} \bar{x} + \mathbb{b}  > \bar{x} > 1 $ and thus violates the constraint.
We observe that for specific choices on the values $ \bar{x} > 1 $ and $ \mathbb{d} > 3 $, we have that $ \bar{y} $ satisfies the required constraints. 
Hence it is possible to find combinations $ \BR{\bar{x}, \, \bar{y}} $ solving the difference equation \eqref{eq:3R-1D2B-S-ISO-EQ-g12-g13-Diff-V2} while satisfying the required constraints. 
The second question now is whether these feasible combinations $ \BR{\bar{x}, \bar{y}} $ are also solutions to the set of equations \eqref{eq:3R-1D2B-S-ISO-EQ-g12-g13-Set}. 
Substituting and rearranging the terms yield
\begin{equation} \label{eq:3R-1D2B-S-ISO-EQ-g12-g13-Diff-Back}
    \frac{1}{1 + \cos \theta^{\star}}\BR{\mathbb{k} \bar{x}^{3} + \mathbb{l} \bar{x}^{2} + \mathbb{m} \bar{x} + \mathbb{n}} = 0
    ,
\end{equation}
where 
\begin{equation} \label{eq:3R-1D2B-S-ISO-EQ-g12-g13-Diff-Back-Coeff}
    \begin{aligned}
        \mathbb{k}
            & = \frac{1}{2} \BR{\mathbb{d} - 2} \BR{\mathbb{d} + 1} \BR{\mathbb{d} - \BR{1 + \cos \theta^{\star}}}
        \\
        \mathbb{l}
            & = - \mathbb{b} \, \mathbb{d} \BR{\mathbb{d} - \BR{1 + \cos \theta^{\star}}}
        \\
        \mathbb{m}
            & = - \frac{1}{2} \BR{\BR{3\mathbb{d} - 2 \cos \theta^{\star}} \BR{\mathbb{d} - 1} + 2}
        \\
        \mathbb{n}
            & = \mathbb{b} \, \mathbb{d}
            .
    \end{aligned}
\end{equation}
With $ \mathbb{d} > 3 $, we obtain $ \mathbb{k} > 0 $, $ \mathbb{l} < 0 $, $ \mathbb{m} < 0 $, and $ \mathbb{n} > 0 $.
Hence we can not provide conclusions on the sign of the cubic term $ \mathbb{k} \bar{x}^{3} + \mathbb{l} \bar{x}^{2} + \mathbb{m} \bar{x} + \mathbb{n} $ on the LHS of \eqref{eq:3R-1D2B-S-ISO-EQ-g12-g13-Diff-Back}. 
In Section \ref{sec:Numerical-Example}, we will numerically evaluate this term. 
Considering the set of constraints $ y > 1 $, $ x < 1 $, $ \mathbb{B} > 0 $, and $ xy -1 > 0 $ in Proposition \ref{prop:3R-1D2B-S-ISO-EQ-Conditions-II}, we would have obtained the same result but with the roles for $ x $ and $ y $ reversed. 

\subsection{Moving configurations}
Consider the equations for the moving configurations and again, let $ \mathbb{A} = x \cos \theta^{\star} - y $ and $ \mathbb{B} = y \cos \theta^{\star} - x $. 
We obtain $ \mathbb{A} \geq 0 \implies \mathbb{B} < 0 $ and $ \mathbb{B} \geq 0 \implies \mathbb{A} < 0 $. 
Since $ x > 0 $, $ y > 0 $, and $ R_{\text{Ad}} > 0 $, it follows that if one of the RHS is non-negative in \eqref{eq:3R-1D2B-S-ISO-MV-g12-g13-Set}, then the RHS of the other equation needs to be negative. 
From Proposition \ref{prop:3R-1D2B-S-ISO-MV-Cubic-Equation-z}, we obtain that the LHS can be negative only when the legs of the isosceles triangle satisfy $ \ell > \widehat{d} $.
In particular, following Corollary \ref{cor:3R-1D2B-S-MV-Cubic-Equation-r}, the cubic equation on the LHS is negative in the range $ \BR{\mathbb{r}_{1}, \, \mathbb{r}_{2}} $.
When $ \ell \leq \widehat{d} $, we do not have feasible combinations $ \BR{x, \, y} $ for the set of equations \eqref{eq:3R-1D2B-S-ISO-MV-g12-g13-Set} and hence the non-existence of moving configurations. 

Assume without loss of generality $ x \in \BR{\mathbb{r}_{1}, \, \mathbb{r}_{2}} $. 
For $ y \leq \mathbb{r}_{1} $, we know the LHS of the second equation in \eqref{eq:3R-1D2B-S-ISO-MV-g12-g13-Set} is positive. 
On the RHS, we require $ \mathbb{B} \geq 0 \implies x < y \implies x < \mathbb{r}_{1} $. 
This contradicts our assumption that $ x \in \BR{\mathbb{r}_{1}, \, \mathbb{r}_{2}} $ and hence we require $ y > \mathbb{r}_{1} $.
Following Corollary \ref{cor:3R-1D2B-S-MV-Cubic-Equation-r}, we can divide $ y > \mathbb{r}_{1} $ in the three regions $ y \in \BR{\mathbb{r}_{1}, \, \mathbb{r}_{2}} $, $ y = \mathbb{r}_{2} $, and $ y > \mathbb{r}_{2} $. 
Before getting into the analysis, we provide the following result concerning the difference equation of \eqref{eq:3R-1D2B-S-ISO-MV-g12-g13-Set}:

\begin{proposition} \label{prop:3R-1D2B-S-ISO-MV-g12-g13-Diff-SOL}
    Let $ x \in \BR{\mathbb{r}_{1}, \, \mathbb{r}_{2}} < 1 $ in \eqref{eq:3R-1D2B-S-ISO-MV-g12-g13-Diff} and $ \mathbb{d} = \frac{1}{2} R_{\text{Ad}} \BR{1 + \cos \theta^{\star}} $.
    We have the following statements:
    \begin{enumerate}
        \item 
        Case: $ y \leq 1 $;
        ~\\
        If $ \mathbb{d} \geq 1 \iff R_{\text{Ad}} \geq \frac{2}{1 + \cos \theta^{\star}} $, then there are no feasible combinations $ \BR{x, \, y} $ that satisfy \eqref{eq:3R-1D2B-S-ISO-MV-g12-g13-Diff}.
        
        \item 
        Case: $ y > 1 $;
        ~\\
        If $ \mathbb{d} \geq 3 \iff R_{\text{Ad}} \geq \frac{6}{1 + \cos \theta^{\star}} $, then the combination $ \BR{\bar{x}, \, \bar{y}} $ where $ \bar{y} = \mathbb{a}\bar{x} + \mathbb{b} $ with $ \mathbb{a} = \frac{\mathbb{d} - 1}{2} $ and $ \mathbb{b} = \frac{1}{2} \sqrt{\BR{\mathbb{d} + 1}\BR{\BR{\mathbb{d} - 3}\bar{x}^{2} + 4}} $ satisfies \eqref{eq:3R-1D2B-S-ISO-MV-g12-g13-Diff}.
    \end{enumerate}
\end{proposition} 

\begin{proof}
    Rearranging the terms in \eqref{eq:3R-1D2B-S-ISO-MV-g12-g13-Diff} yields 
    \begin{equation} \label{eq:3R-1D2B-S-ISO-MV-g12-g13-Diff-V2}
        \begin{aligned}
            x^{2} + y^{2} + \BR{1 - \mathbb{d}} xy 
                = 1 + \mathbb{d}
                .
        \end{aligned}
    \end{equation}
    
    \begin{enumerate}
        \item
        Case: $ y \leq 1 $;
        ~\\
        Choosing $ \mathbb{d} \geq 1 $, we have $ \BR{1 - \mathbb{d}} \leq 0 $. 
        The LHS of \eqref{eq:3R-1D2B-S-ISO-MV-g12-g13-Diff-V2} has upper bound $ x^{2} + y^{2} + \BR{1 - \mathbb{d}} xy \leq x^{2} + y^{2} < 2 $ while the RHS has a value
        $ \BR{1 + \mathbb{d}} \geq 2 $.
        We infer $ \mathbb{d} \geq 1 \iff  R_{\text{Ad}} \geq \frac{2}{1 + \cos \theta^{\star}} $ yields no solution for \eqref{eq:3R-1D2B-S-ISO-MV-g12-g13-Diff-V2}. 
        
        \item 
        Case: $ y > 1 $;
        ~\\
        We consider a specific value for $ x = \bar{x} $ in the given range. 
        Note that $ \bar{x} < 1 $. 
        Solving \eqref{eq:3R-1D2B-S-ISO-MV-g12-g13-Diff-V2} for the unknown $ y $, we obtain 
        \begin{equation} \label{eq:3R-1D2B-S-ISO-MV-g12-g13-Diff-Roots}
            y = \frac{\mathbb{d} - 1}{2} \bar{x} \pm \frac{1}{2} \sqrt{\BR{\mathbb{d} + 1} \BR{\BR{\mathbb{d} - 3} \bar{x}^{2} + 4}}
            .
        \end{equation}
        With $ \mathbb{d} \geq 3 $, we have that the term under the square root is positive. 
        Also, $ \frac{\mathbb{d} - 1}{2} \geq 1 $.
        Applying Descartes' rule of signs, we infer that for $ \mathbb{d} \geq 3 $, we have one positive and one negative root. 
        The positive root is then $ \bar{y} = \frac{\mathbb{d} - 1}{2} \bar{x} + \frac{1}{2} \sqrt{\BR{\mathbb{d} + 1} \BR{\BR{\mathbb{d} - 3} \bar{x}^{2} + 4}} $.
    \end{enumerate}
    This completes the proof. 
\end{proof}

For moving configurations, we state the following result:

\begin{lemma} \label{lem:} \label{lem:3R-1D2B-S-ISO-MV-Main-Acute}
    Consider a team of three robots moving according to \eqref{eq:3R-1D2B-S-p-Dynamics} with specific gains $ K_{\text{d}} > 0 $ and $ K_{\text{b}} > 0 $.
    Define the gain ratios $ R_{\text{bd}} = \frac{K_{\text{b}}}{K_{\text{d}}} $ and $ R_{\text{Ad}} = \frac{K_{\text{A}}}{K_{\text{d}}} $ with $ K_{\text{A}} > 0 $, and $ \widehat{d} = \sqrt{3} \sqrt[3]{\frac{R_{\text{bd}}}{2}} $. 
    Furthermore, let the desired formation shape be an isosceles triangle with legs $ \ell > \widehat{d} $ and vertex angle $ \theta^{\star} \in \BR{0^{\degree}, \, 90^{\degree}} $.
    Finally, assume $ x \in \BR{\mathbb{r}_{1}, \, \mathbb{r}_{2}} $ and $ y > \mathbb{r}_{1} $, where $ x $ and $ y $ are defined in \eqref{eq:3R-1D2B-S-ISO-EQ-PAR-XY}.
    If $ R_{\text{Ad}} \geq \max \CBR{\frac{6}{1 + \cos \theta^{\star}}, \, \frac{2}{1 - \cos \theta^{\star}}} $, then there are no feasible combinations $ \BR{x, \, y} $ satisfying the set of equations \eqref{eq:3R-1D2B-S-ISO-MV-g12-g13-Set}. 
\end{lemma}

\begin{proof}
    The proof will be given for the three regions of $ y > \mathbb{r}_{1} $ obtained from Corollary \ref{cor:3R-1D2B-S-MV-Cubic-Equation-r}:
    \begin{enumerate}
        \item 
        $ x \in \BR{\mathbb{r}_{1}, \, \mathbb{r}_{2}} $ and $ y \in \BR{\mathbb{r}_{1}, \, \mathbb{r}_{2}} $;
        ~\\
        In this region, we can divide the analysis to the cases $ x = y $ and $ x \neq y $:
        \begin{enumerate}
            \item 
            Case: $ x = y $;
            ~\\
            Rearranging the terms in \eqref{eq:3R-1D2B-S-ISO-MV-g12-g13-Equal} yields 
            \begin{equation} \label{eq:3R-1D2B-S-ISO-MV-g12-g13-Equal-Acute}
                \begin{aligned}
                    & 
                    x \ell^{3} \BR{\BR{\mathbb{c} + 1} x^{2} + \BR{\mathbb{c} - 1}} + R_{\text{bd}}
                        = 0
                    , 
                \end{aligned}
            \end{equation}
            where $ \mathbb{c} = \frac{1}{2} R_{\text{Ad}} \BR{1 - \cos \theta^{\star}} $.
            Choosing $ \mathbb{c} \geq 1 $, we have $ \BR{\mathbb{c} + 1} \geq 2 $ and $ \BR{\mathbb{c} - 1} \geq 0 $.
            Therefore, all terms in \eqref{eq:3R-1D2B-S-ISO-MV-g12-g13-Equal-Acute} are non-negative and at least one term is positive; then the sum on the LHS is also positive.
            The equation \eqref{eq:3R-1D2B-S-ISO-MV-g12-g13-Equal-Acute} does not have roots for $ \mathbb{c} \geq 1 \iff R_{\text{Ad}} \geq \frac{2}{1 - \cos \theta^{\star}} $.
            
            \item 
            Case: $ x \neq y $;
            ~\\
            Since both $ x < 1 $ and $ y < 1 $, it follows from Proposition \ref{prop:3R-1D2B-S-ISO-MV-g12-g13-Diff-SOL} that the difference equation \eqref{eq:3R-1D2B-S-ISO-MV-g12-g13-Diff} does not have a solution when the gain ratio $ R_{\text{Ad}} $ satisfies $ R_{\text{Ad}} \geq \frac{2}{1 + \cos \theta^{\star}} $, 
            This implies that the set of equations \eqref{eq:3R-1D2B-S-ISO-MV-g12-g13-Set} also does not have a solution.
        \end{enumerate}
        Combining both cases, we infer that $ R_{\text{Ad}} \geq \frac{2}{1 - \cos \theta^{\star}} $ will yield no feasible combinations $ \BR{x, \, y} $ for \eqref{eq:3R-1D2B-S-ISO-MV-g12-g13-Set}.
        
        \item 
        $ x \in \BR{\mathbb{r}_{1}, \, \mathbb{r}_{2}} $ and $ y = \mathbb{r}_{2} $;
        ~\\
        We have $ x < 1 $ and $ y = \mathbb{r}_{2} < 1 $.
        From Proposition \ref{prop:3R-1D2B-S-ISO-MV-g12-g13-Diff-SOL}, we infer that when $ R_{\text{Ad}} \geq \frac{2}{1 + \cos \theta^{\star}} $, we have no feasible combinations $ \BR{x, \, y} $ for \eqref{eq:3R-1D2B-S-ISO-MV-g12-g13-Set}.
        
        \item
        $ x \in \BR{\mathbb{r}_{1}, \, \mathbb{r}_{2}} $ and $ y > \mathbb{r}_{2} $;
        ~\\
        We can divide the region for $ y $ in two sub-regions, namely $ \mathbb{r_{2}} < y \leq 1 $ and $ y > 1 $. 
        \begin{enumerate}
            \item 
            Case: $ y \in \left(\mathbb{r}_{2}, \, 1\right] $;
            ~\\
            Since $ x < 1 $ and $ y \leq 1 $, following Proposition \ref{prop:3R-1D2B-S-ISO-MV-g12-g13-Diff-SOL}, we infer $ R_{\text{Ad}} \geq \frac{2}{1 + \cos \theta^{\star}} $ is sufficient to obtain no feasible combinations $ \BR{x, \, y} $ for \eqref{eq:3R-1D2B-S-ISO-MV-g12-g13-Set}.
            
            \item 
            Case: $ y > 1 $;
            ~\\
            With the particular choice $ R_{\text{Ad}} \geq \frac{6}{1 + \cos \theta^{\star}} \iff \mathbb{d} \geq 3 $ from Proposition \ref{prop:3R-1D2B-S-ISO-MV-g12-g13-Diff-SOL}, we obtain $ \bar{y} = \mathbb{a}\bar{x} + \mathbb{b} $ with $ \mathbb{a} = \frac{\mathbb{d} - 1}{2} $ and $ \mathbb{b} = \frac{1}{2} \sqrt{\BR{\mathbb{d} + 1} \BR{\BR{\mathbb{d} - 3} \bar{x}^{2} + 4}} $ is the solution to \eqref{eq:3R-1D2B-S-ISO-MV-g12-g13-Diff} for a fixed $ \bar{x} $. 
            Substituting the obtained pair $ \BR{\bar{x}, \, \bar{y}} $ back in \eqref{eq:3R-1D2B-S-ISO-MV-g12-g13-Set} yields
            \begin{equation} \label{eq:3R-1D2B-S-ISO-MV-g12-g13-Diff-Back}
                \frac{1}{1 + \cos \theta^{\star}}\BR{\mathbb{k} \bar{x}^{3} + \mathbb{l} \bar{x}^{2} + \mathbb{m} \bar{x} + \mathbb{n}} \ell^{3} + R_{\text{bd}} = 0
                ,
            \end{equation}
            where 
            \begin{equation} \label{eq:3R-1D2B-S-ISO-MV-g12-g13-Diff-Back-Coeff}
                \begin{aligned}
                    \mathbb{k}
                        & = \frac{1}{2} \BR{\mathbb{d} - 2} \BR{\mathbb{d} + 1} \BR{\mathbb{d} - \BR{1 + \cos \theta^{\star}}}
                    \\
                    \mathbb{l}
                        & = \mathbb{b} \, \mathbb{d} \BR{\mathbb{d} - \BR{1 + \cos \theta^{\star}}}
                    \\
                    \mathbb{m}
                        & = \frac{1}{2} \BR{\mathbb{d} + 1} \BR{3 \mathbb{d} - 2\BR{1 + \cos \theta^{\star}}}
                    \\
                    \mathbb{n}
                        & = \mathbb{b} \, \mathbb{d}
                \end{aligned}
            \end{equation}
            With $ \mathbb{d} \geq 3 $, we obtain $ \mathbb{k} > 0 $, $ \mathbb{l} > 0 $, $ \mathbb{m} > 0 $, and $ \mathbb{n} > 0 $.
            Since the individual terms are positive, it follows the sum is also positive; the combination $ \BR{\bar{x}, \, \bar{y}} $ satisfying \eqref{eq:3R-1D2B-S-ISO-MV-g12-g13-Diff} is not a solution to \eqref{eq:3R-1D2B-S-ISO-MV-g12-g13-Set}.
            So solutions to \eqref{eq:3R-1D2B-S-ISO-MV-g12-g13-Set} are of the form $ \BR{x, \, y} \neq \BR{\bar{x}, \, \bar{y}} $. 
            However, these will not solve \eqref{eq:3R-1D2B-S-ISO-MV-g12-g13-Diff} and therefore, we conclude that for $ R_{\text{Ad}} \geq \frac{6}{1 + \cos \theta^{\star}} $, the solution set to \eqref{eq:3R-1D2B-S-ISO-MV-g12-g13-Set} is empty for the mentioned region of $ x $ and $ y $. 
        \end{enumerate}
        Combining the results of both parts, we obtain that $ R_{\text{Ad}} \geq \frac{6}{1 + \cos \theta^{\star}} $ is sufficient to obtain no feasible combinations $ \BR{x, \, y} $ for \eqref{eq:3R-1D2B-S-ISO-MV-g12-g13-Set}.
    \end{enumerate}
    Gathering the results for all the three considered regions for $ y $, we conclude that $ R_{\text{Ad}} \geq \max \CBR{\frac{6}{1 + \cos \theta^{\star}}, \, \frac{2}{1 - \cos \theta^{\star}}} $.
    This completes the proof. 
\end{proof}

From a design perspective, provided $ \ell $ is given, we can tune the gains $ K_{\text{d}} $ and $ K_{\text{b}} $ such that $ \ell \leq \widehat{d} $ is satisfied. 
In this scenario, any choice of $ K_{\text{A}} > 0 $ would yield no solutions $ \BR{x, \, y} $ to the set of equations \eqref{eq:3R-1D2B-S-ISO-MV-g12-g13-Set}.
This is in agreement with our earlier work \cite{Chan2020} in which moving configurations may occur only when $ d_{12}^{\star} \geq \widehat{d} $ and $ d_{13}^{\star} \geq \widehat{d} $ hold.
We need Lemma \ref{lem:3R-1D2B-S-ISO-MV-Main-Acute} when tuning the gains $ K_{\text{d}} $ and $ K_{\text{b}} $ only is not enough. 
It provides a lower bound on the gain $ K_{\text{A}} $ for a chosen $ K_{\text{d}} $.
In the region $ \theta^{\star} \in \left[60^{\degree}, \, 90^{\degree} \right) $, this lower bound is $ K_{\text{A}} \geq \frac{6}{1 + \cos \theta^{\star}} K_{\text{d}} $ while $ K_{\text{A}} \geq \frac{2}{1 - \cos \theta^{\star}} K_{\text{d}}$ when the vertex angle is $ \theta^{\star} \in \BR{0, \, 60} $ in Lemma \ref{lem:3R-1D2B-S-ISO-MV-Main-Acute}.
Furthermore, Proposition \ref{prop:3R-1D2B-S-ISO-MV-g12-g13-Diff-SOL} and Lemma \ref{lem:3R-1D2B-S-ISO-MV-Main-Acute} are results which hold for $ x \in \BR{\mathbb{r}_{1}, \, \mathbb{r}_{2}} $ and $ y > \mathbb{r}_{1} $. 
For the case $ y \in \BR{\mathbb{r}_{1}, \, \mathbb{r}_{2}} $ and $ x > \mathbb{r}_{1} $, we obtain the same result, albeit the roles of $ x $ and $ y $ are reversed.
We conclude this section with the following main result:

\begin{theorem} \label{thm:3R-1D2B-S-Global-Convergence-Acute}
    Consider a team of three robots moving according to \eqref{eq:3R-1D2B-S-p-Dynamics} with $ K_{\text{d}} > 0 $, $ K_{\text{b}} > 0 $, and $ K_{\text{A}} > 0 $. 
    Define the gain ratios $ R_{\text{bd}} = \frac{K_{\text{b}}}{K_{\text{d}}} $ and $ R_{\text{Ad}} = \frac{K_{\text{A}}}{K_{\text{d}}} $, and also $ \widehat{d} = \sqrt{3} \sqrt[3]{\frac{R_{\text{bd}}}{2}} $. 
    Let the desired formation shape be an acute isosceles triangle with legs $ \ell > 0 $ and vertex angle $ \theta^{\star} $. 
    Finally, parametrize the distances $ d_{12} $ and $ d_{13} $ as in \eqref{eq:3R-1D2B-S-ISO-EQ-PAR-XY}.
    Then, starting from all feasible initial configurations, the robots converge to a desired equilibrium configuration $ p_{\text{eq}} \in \mathcal{S}_{p} $ in which all the individual tasks are attained if either $ \BR{K_{\text{d}}, \, K_{\text{b}}, \, K_{\text{A}}} $ is chosen such that $ \ell \leq \widehat{d} $ and $ R_{\text{Ad}} \leq \frac{6}{1 + \cos \theta^{\star}} $ for $  \theta^{\star} \in \BR{0^{\degree}, \, 90^{\degree}} $ or if $ \BR{K_{\text{d}}, \, K_{\text{b}}, \, K_{\text{A}}} $ is chosen such that $ \ell > \widehat{d} $ and $ R_{\text{Ad}} = \frac{6}{1 + \cos \theta^{\star}} $ for $ \theta^{\star} \in \left[60^{\degree}, \, 90^{\degree} \right) $. 
\end{theorem}

\begin{proof}
    The proof follows directly from the results obtained in Lemmas \ref{lem:3R-1D2B-S-ISO-EQ-Main-Acute} and \ref{lem:3R-1D2B-S-ISO-MV-Main-Acute}.
    This completes the proof.
\end{proof}

With the current constraints imposed on the gain ratio $ R_{\text{Ad}} $, we can not provide convergence results for isosceles triangles with legs $ \ell > \widehat{d} $ and vertex angle $ \theta^{\star} \in \BR{0^{\degree}, \, 60^{\degree}} $.
We will numerically evaluate this in Section \ref{sec:Numerical-Example}.


\section{Right and Obtuse Isosceles Triangles} \label{sec:Isosceles-Triangle-Analysis-Right-Obtuse}
We continue with right and obtuse isosceles triangles in which the vertex angle $ \theta^{\star} $ is in the range $ \theta^{\star} \in \left[90^{\degree}, \, 180^{\degree}\right) \iff \cos \theta^{\star} \in \left(-1, \, 0\right] $.

\subsection{Equilibrium configurations}
We state the following result on equilibrium configurations for right and obtuse isosceles triangles:

\begin{lemma} \label{lem:3R-1D2B-S-ISO-EQ-Main-Right-Obtuse}
    Consider a team of three robots moving according to \eqref{eq:3R-1D2B-S-p-Dynamics} with gains $ K_{\text{d}} > 0 $, $ K_{\text{b}} > 0 $, and $ K_{\text{A}} > 0 $. 
    Let the desired formation shape be an isosceles triangle with legs $ \ell > 0 $ and vertex angle $ \theta^{\star} \in \left[90^{\degree}, \, 180^{\degree}\right) $.
    In addition, parametrize the distances $ d_{12} $ and $ d_{13} $ as in \eqref{eq:3R-1D2B-S-ISO-EQ-PAR-XY}.
    Then the equilibrium configurations $ p_{\text{eq}} $ are all in the desired set $ \mathcal{S}_{p} $.
\end{lemma}

\begin{proof}
    Given $ \theta^{\star} \in \left[90^{\degree}, \, 180^{\degree}\right) $, we obtain $ \cos \theta^{\star} \in \left(-1, \, 0\right] $.
    Correspondingly, in \eqref{eq:3R-1D2B-S-ISO-EQ-g12-g13-Set}, $ \mathbb{A} = x \cos \theta^{\star} - y \leq -y < 0 $ and $ \mathbb{B} = y \cos \theta^{\star} - x \leq -x < 0 $.
    Assuming $ xy - 1 > 0 $ on the RHS leads to $ x < 1 $ and $ y < 1 $ on the LHS; in turn, this results in $ xy < 1 $ and therefore contradicting the assumption. 
    Similar arguments hold for $ xy - 1 < 0 $. 
    Hence there are no feasible combinations $ \BR{x, \, y} $ satisfying $ xy - 1 \neq 0 $.
    In addition, it follows from Proposition \ref{prop:3R-1D2B-S-ISO-EQ-Conditions-I} that $ xy - 1 = 0 $ is equivalent to the combination $ \BR{x, \, y} = \BR{1, \, 1} $; Robot \texttt{R1} attains all its assigned tasks. 
    With robots \texttt{R2} and \texttt{R3} also attaining its individual task, we conclude that all robots in the team attain its individual tasks, i.e., $ e = \mathbb{0}_{7} $.
    This completes the proof.
\end{proof}

In Lemma \ref{lem:3R-1D2B-S-ISO-EQ-Main-Right-Obtuse}, we do not have to impose additional constraints on the gains $ K_{\text{d}} $, $ K_{\text{b}} $, and $ K_{\text{A}} $ other than that they should be positive. 

\subsection{Moving configurations}
For moving configurations, we observe that the sign on the RHS of \eqref{eq:3R-1D2B-S-ISO-MV-g12-g13-Set} depends only on the terms $ \mathbb{A} = x \cos \theta^{\star} - y $ and $ \mathbb{B} = y \cos \theta^{\star} - x $. 
Since $ \cos \theta^{\star} \in \left(-1, \, 0\right] $, it follows that $ \mathbb{A} < 0 $ and $ \mathbb{B} < 0 $ implying the RHS of \eqref{eq:3R-1D2B-S-ISO-MV-g12-g13-Set} is negative. 
For the LHS to be also negative, we require from Proposition \ref{prop:3R-1D2B-S-ISO-MV-Cubic-Equation-z} that the desired length $ \ell $ should satisfy $ \ell > \widehat{d} $.
In particular, the LHS is negative when $ x \in \BR{\mathbb{r}_{1}, \, \mathbb{r}_{2}} $ and $ y \in \BR{\mathbb{r}_{1}, \, \mathbb{r}_{2}} $ with $ \mathbb{r}_{1} $ and $ \mathbb{r}_{2} $ given in Corollary \ref{cor:3R-1D2B-S-MV-Cubic-Equation-r}.
The following lemma states a condition on $ R_{\text{Ad}} $ for precluding moving configurations when $ \ell > \widehat{d} $:

\begin{lemma} \label{lem:3R-1D2B-S-ISO-MV-Main-Right-Obtuse}
    Consider a team of three robots moving according to \eqref{eq:3R-1D2B-S-p-Dynamics} with specific gains $ K_{\text{d}} > 0 $ and $ K_{\text{b}} > 0 $.
    Define the gain ratios $ R_{\text{bd}} = \frac{K_{\text{b}}}{K_{\text{d}}} $ and $ R_{\text{Ad}} = \frac{K_{\text{A}}}{K_{\text{d}}} $ with $ K_{\text{A}} > 0 $, and also $ \widehat{d} = \sqrt{3} \sqrt[3]{\frac{R_{\text{bd}}}{2}} $.
    Furthermore, let the desired formation shape be an isosceles triangle with legs $ \ell > \widehat{d} $ and vertex angle $ \theta^{\star} \in \left[90^{\degree}, \, 180^{\degree}\right) $.
    Finally, parametrize the distances $ d_{12} $ and $ d_{13} $ as in \eqref{eq:3R-1D2B-S-ISO-EQ-PAR-XY}.
    If $ R_{\text{Ad}} \geq \frac{2}{1 + \cos \theta^{\star}} $, then there are no feasible combinations $ \BR{x, \, y} $ satisfying the set of equations \eqref{eq:3R-1D2B-S-ISO-MV-g12-g13-Set}.
\end{lemma}

\begin{proof}
    The proof is divided in two parts, namely considering $ x = y $ and $ x \neq y $ in the feasible region $ \BR{\mathbb{r}_{1}, \, \mathbb{r}_{2}}^{2} $.
    \begin{enumerate}
        \item 
        Case: $ x = y $;
        ~\\
        Rearranging the terms in \eqref{eq:3R-1D2B-S-ISO-MV-g12-g13-Equal} yields 
        \begin{equation} \label{eq:3R-1D2B-S-ISO-MV-g12-g13-Equal-Right-Obtuse}
            \begin{aligned}
                & 
                x \ell^{3} \BR{\BR{\mathbb{c} + 1} x^{2} + \BR{\mathbb{c} - 1}} + R_{\text{bd}}
                    = 0
                , 
            \end{aligned}
        \end{equation}
        where $ \mathbb{c} = \frac{1}{2} R_{\text{Ad}} \BR{1 - \cos \theta^{\star}} $.
        Choosing $ \mathbb{c} \geq 1 $, we have $ \BR{\mathbb{c} + 1} \geq 2 $ and $ \BR{\mathbb{c} - 1} \geq 0 $.
        Therefore, all terms in \eqref{eq:3R-1D2B-S-ISO-MV-g12-g13-Equal-Right-Obtuse} are non-negative and at least one term is positive; then, the sum on the LHS is also positive.
        The equation \eqref{eq:3R-1D2B-S-ISO-MV-g12-g13-Equal-Right-Obtuse} does not have roots for $ \mathbb{c} \geq 1 \iff R_{\text{Ad}} \geq \frac{2}{1 - \cos \theta^{\star}} $.
        
        \item 
        Case: $ x \neq y $;
        ~\\
        It follows from Proposition \ref{prop:3R-1D2B-S-ISO-MV-g12-g13-Diff-SOL} that \eqref{eq:3R-1D2B-S-ISO-MV-g12-g13-Diff} does not have a solution for $ R_{\text{Ad}} \geq \frac{2}{1 + \cos \theta^{\star}} $ since $ x < 1 $ and $ y < $ 1.
        This in turn implies the set of equations \eqref{eq:3R-1D2B-S-ISO-MV-g12-g13-Set} does not have a solution, since solutions to \eqref{eq:3R-1D2B-S-ISO-MV-g12-g13-Set} are immediate solutions to \eqref{eq:3R-1D2B-S-ISO-MV-g12-g13-Diff}.
    \end{enumerate}
    We have obtained two lower bounds on $ R_{\text{Ad}} $ for the different cases. 
    Notice that $ \frac{2}{1 + \cos \theta^{\star}} \geq \frac{2}{1 - \cos \theta^{\star}} $ for $ \cos \theta^{\star} \leq 0 $.
    Hence the choice of $ R_{\text{Ad}} \geq \frac{2}{1 + \cos \theta^{\star}} $ will yield no feasible combinations $ \BR{x, \, y} $ satisfying the set of equations  \eqref{eq:3R-1D2B-S-ISO-MV-g12-g13-Set}.
    This completes the proof.
\end{proof}

Similar to acute isosceles triangles, provided $ \ell $ is given, we can first tune the gains $ K_{\text{d}} $ and $ K_{\text{b}} $ such that $ \ell \leq \widehat{d} $ is satisfied. 
Then any choice of $ K_{\text{A}} > 0 $ would yield no solutions $ \BR{x, \, y} $ to the set of equations \eqref{eq:3R-1D2B-S-ISO-MV-g12-g13-Set}.
We need Lemma \ref{lem:3R-1D2B-S-ISO-MV-Main-Right-Obtuse} when tuning the gains $ K_{\text{d}} $ and $ K_{\text{b}} $ only is not enough. 
In that case, for a specific value $ K_{\text{d}} > 0 $, we know from the lemma that the gain for the signed area control term needs to satisfy $ K_{\text{A}} \geq \frac{2}{1 + \cos \theta^{\star}} K_{\text{d}} $.
When $ \cos \theta^{\star} \to -1 $, i.e., when the vertex angle $ \theta^{\star} $ of the isosceles triangle is close to $ 180^{\degree} $, we obtain that $ K_{\text{A}} \to \infty $.

Combining the analyses on the equilibrium and moving configurations, we have the following result for the three-robot formation tasked with displaying a right or obtuse isosceles triangle with legs $ \ell > 0 $:

\begin{theorem} \label{thm:3R-1D2B-S-Global-Convergence-Right-Obtuse}
    Consider a team of three robots moving according to \eqref{eq:3R-1D2B-S-p-Dynamics} with gains $ K_{\text{d}} > 0 $, $ K_{\text{b}} > 0 $, and $ K_{\text{A}} > 0 $.
    Define the gain ratios $ R_{\text{bd}} = \frac{K_{\text{b}}}{K_{\text{d}}} $ and $ R_{\text{Ad}} = \frac{K_{\text{A}}}{K_{\text{d}}} $, and also $ \widehat{d} = \sqrt{3} \sqrt[3]{\frac{R_{\text{bd}}}{2}} $. 
    Furthermore, let the desired formation shape be an isosceles triangle with legs $ \ell > 0 $ and vertex angle $ \theta^{\star} \in \left[90^{\degree}, \, 180^{\degree}\right) $. 
    Finally, parametrize the distances $ d_{12} $ and $ d_{13} $ as in \eqref{eq:3R-1D2B-S-ISO-EQ-PAR-XY}.
    Then, starting from all feasible initial configurations, the robots converge to a desired equilibrium configuration $ p_{\text{eq}} \in \mathcal{S}_{p} $ if $ \BR{K_{\text{d}}, \, K_{\text{b}}} $ is chosen such that $ \ell \leq \widehat{d} $ or if $ \BR{K_{\text{d}}, \, K_{\text{b}}, \, K_{\text{A}}} $ is chosen such that $ \ell > \widehat{d} $ and $ R_{\text{Ad}} \geq \frac{2}{1 + \cos \theta^{\star}} $. 
\end{theorem}

\begin{proof}
    The proof follows directly from the results obtained in Lemmas \ref{lem:3R-1D2B-S-ISO-EQ-Main-Right-Obtuse} and \ref{lem:3R-1D2B-S-ISO-MV-Main-Right-Obtuse}.
    This completes the proof.
\end{proof}

In Table \ref{tab:3R-1D2B-S-ISO-Summary-Results}, we summarize the results on the gain ratio $ R_{\text{Ad}} $ such that convergence to the desired isosceles triangular formation is obtained. 
\begin{table*}[!tb]
    \centering
    \caption{Conditions on the gain ratio $ R_{\text{Ad}} = \frac{K_{\text{A}}}{K_{\text{d}}} $ in Theorems \ref{thm:3R-1D2B-S-Global-Convergence-Acute} and \ref{thm:3R-1D2B-S-Global-Convergence-Right-Obtuse} and Corollary \ref{cor:3R-1D2B-S-Global-Convergence-Acute-V2} for yielding convergence to the desired isosceles triangular formation.}
    \label{tab:3R-1D2B-S-ISO-Summary-Results}
    \begin{tabularx}{0.975\linewidth}{llXl}
        \toprule
            & Acute isosceles triangle
            & Acute isosceles triangle (with numerics)
            & Right \& obtuse isosceles triangle
            \\
        \midrule
        \addlinespace
        $ \ell \leq \widehat{d} $
            & $ R_{\text{Ad}} \in \left(0, \, \frac{6}{1 + \cos \theta^{\star}} \right] $
            & $ R_{\text{Ad}} \in \BR{0, \infty} $
            & $ R_{\text{Ad}} \in \BR{0, \infty} $
            \\
        \addlinespace
        $ \ell > \widehat{d} $
            & $ R_{\text{Ad}} = \frac{6}{1 + \cos \theta^{\star}} $ for $ \theta^{\star} \in \left[60^{\degree}, \, 90^{\degree} \right) $ 
            & $ R_{\text{Ad}} \in \left[ \max \CBR{\frac{6}{1 + \cos \theta^{\star}}, \frac{2}{1 - \cos \theta^{\star}}}, \, \infty \right) $
            & $ R_{\text{Ad}} \in \left[ \frac{2}{1 + \cos \theta^{\star}}, \, \infty \right) $
            \\
        \bottomrule
    \end{tabularx}
\end{table*}

\section{Numerical Example} \label{sec:Numerical-Example}

\subsection{Numerical evaluation of \eqref{eq:3R-1D2B-S-ISO-EQ-g12-g13-Diff-Back}}
\begin{figure*}[!tb]
    \centering
    {
    \subfigure[$ \theta^{\star} = 5^{\degree} $ ]
    {
    \includegraphics[width=0.475\textwidth]{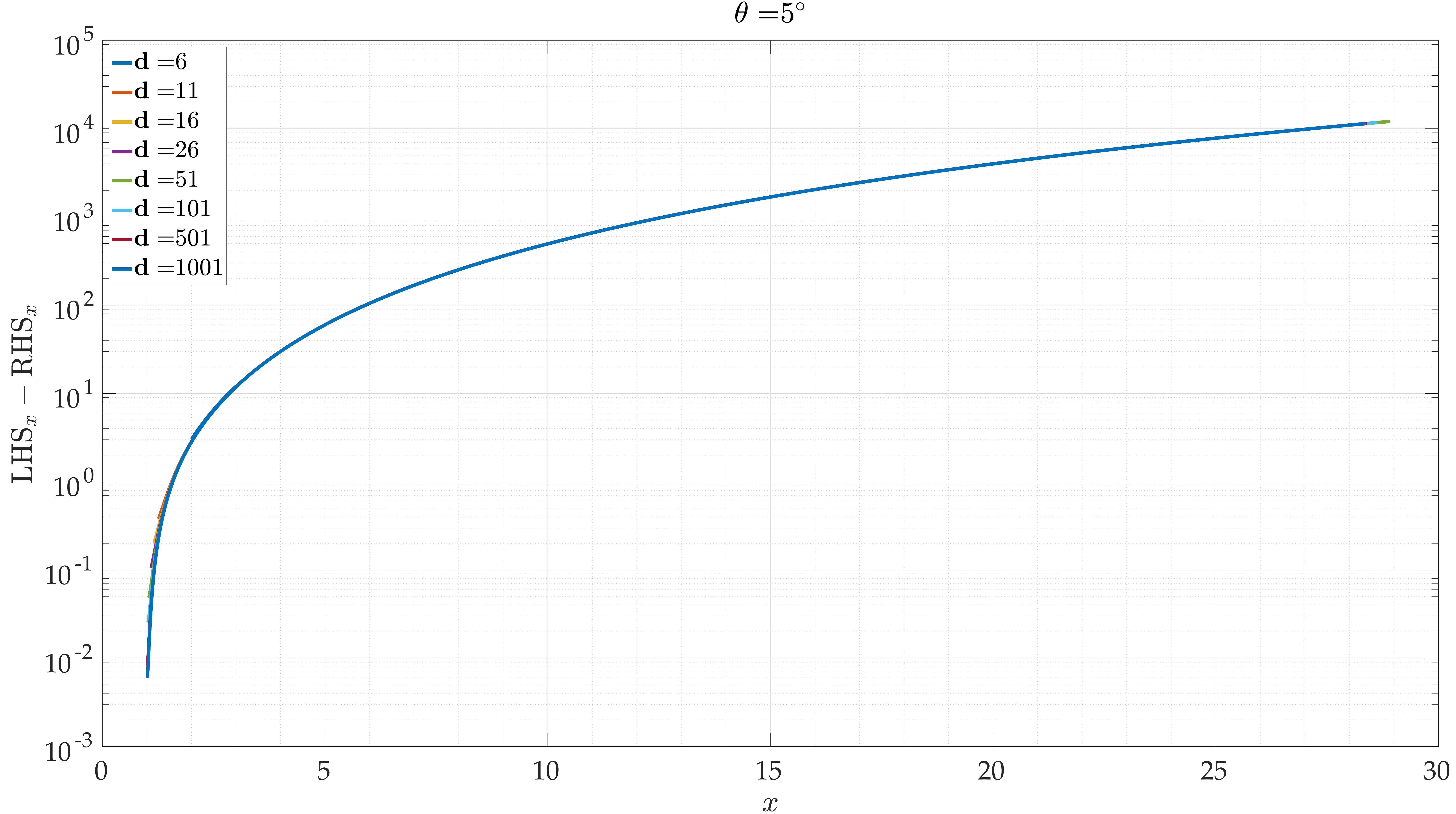}
    \label{fig:3R-1D2B-S-ISO-EQ-NUM-theta-005}
    }
    \hfill
    \subfigure[$ \theta^{\star} = 15^{\degree} $ ]
    {\includegraphics[width=0.475\textwidth]{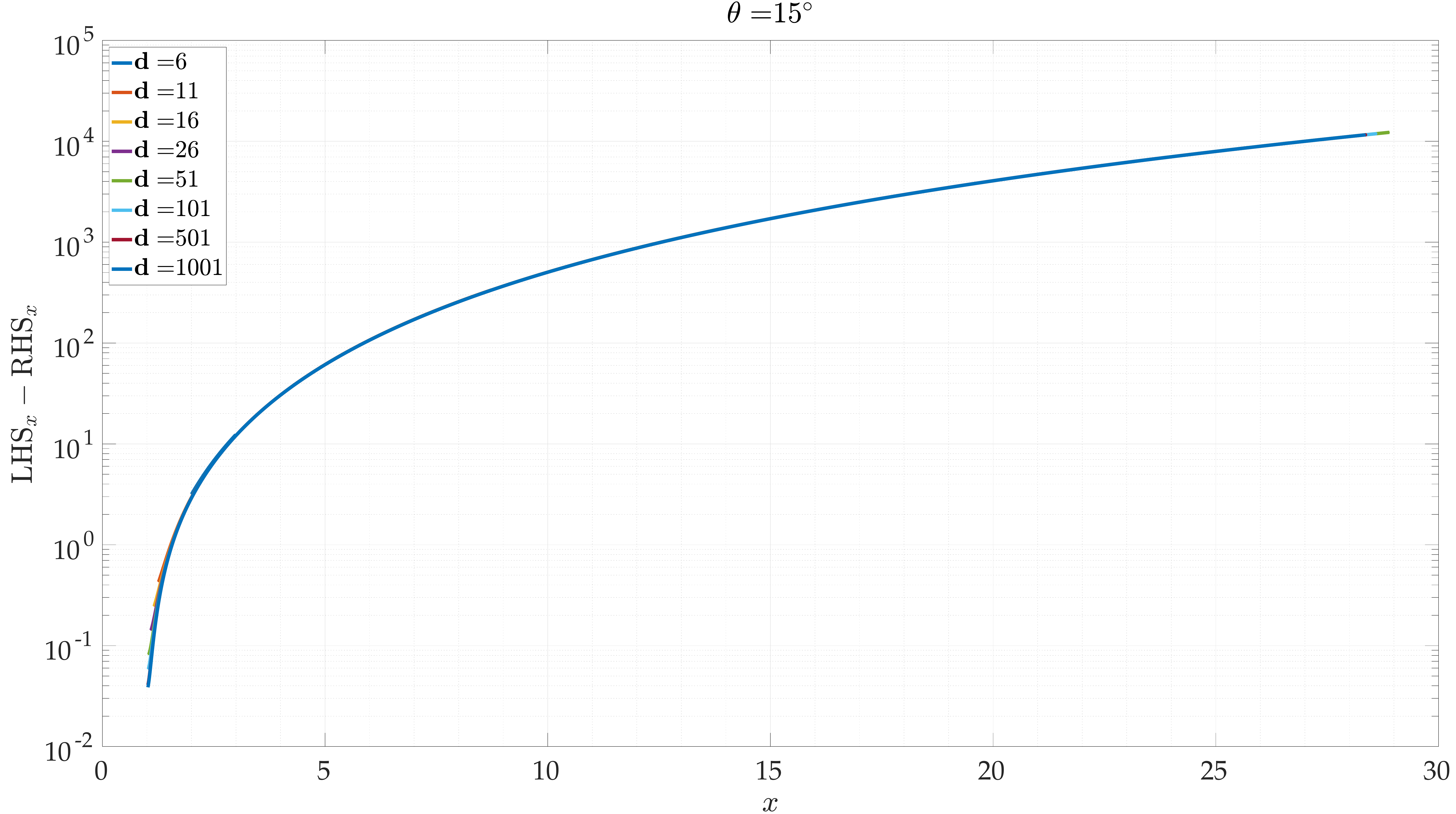}
    \label{fig:3R-1D2B-S-ISO-EQ-NUM-theta-015}
    }
    \hfill
    \subfigure[$ \theta^{\star} = 45^{\degree} $ ]
    {
    \includegraphics[width=0.475\textwidth]{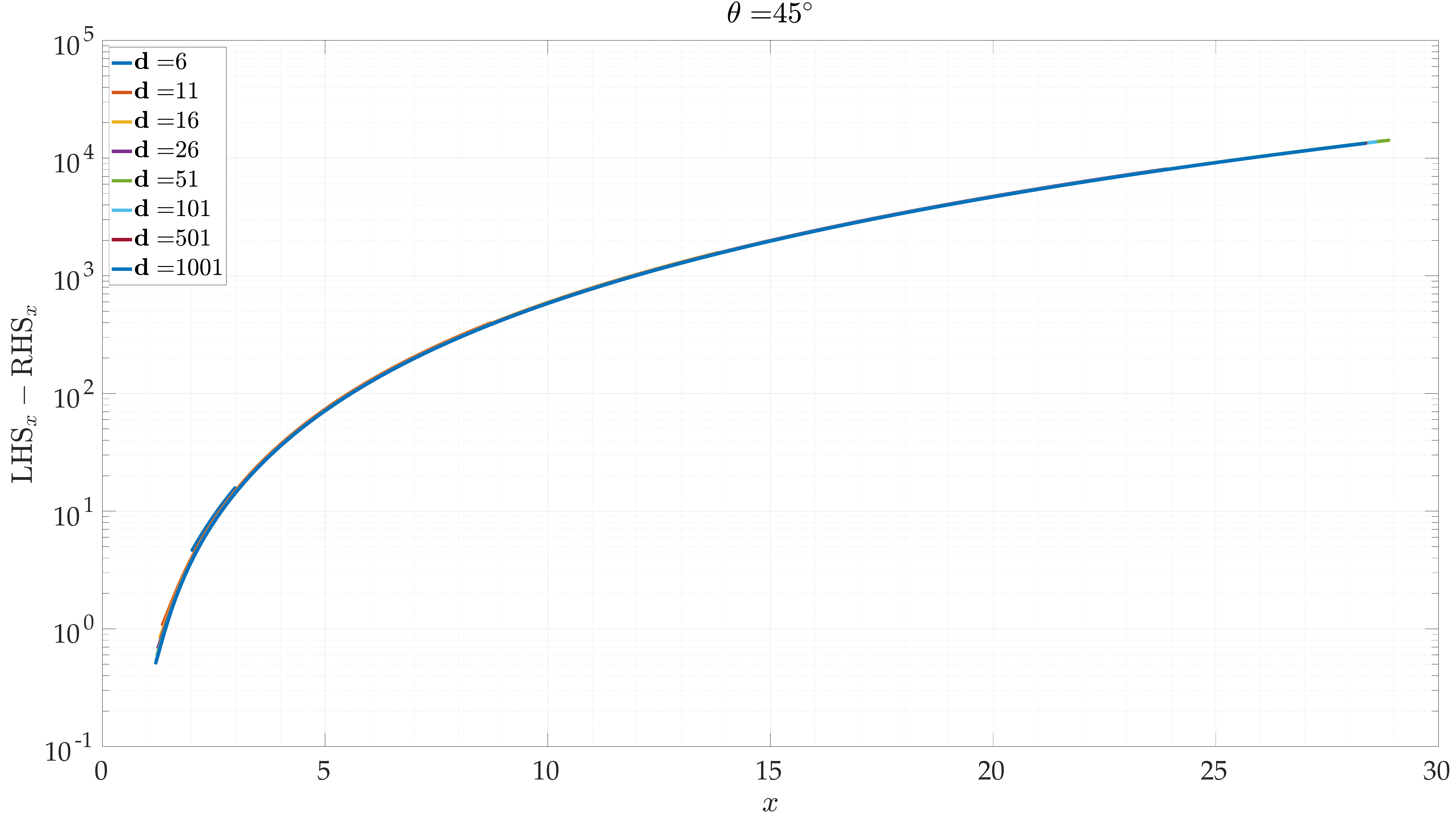}
    \label{fig:3R-1D2B-S-ISO-EQ-NUM-theta-045}
    }
    \hfill
    \subfigure[$ \theta^{\star} = 75^{\degree} $ ]
    {\includegraphics[width=0.475\textwidth]{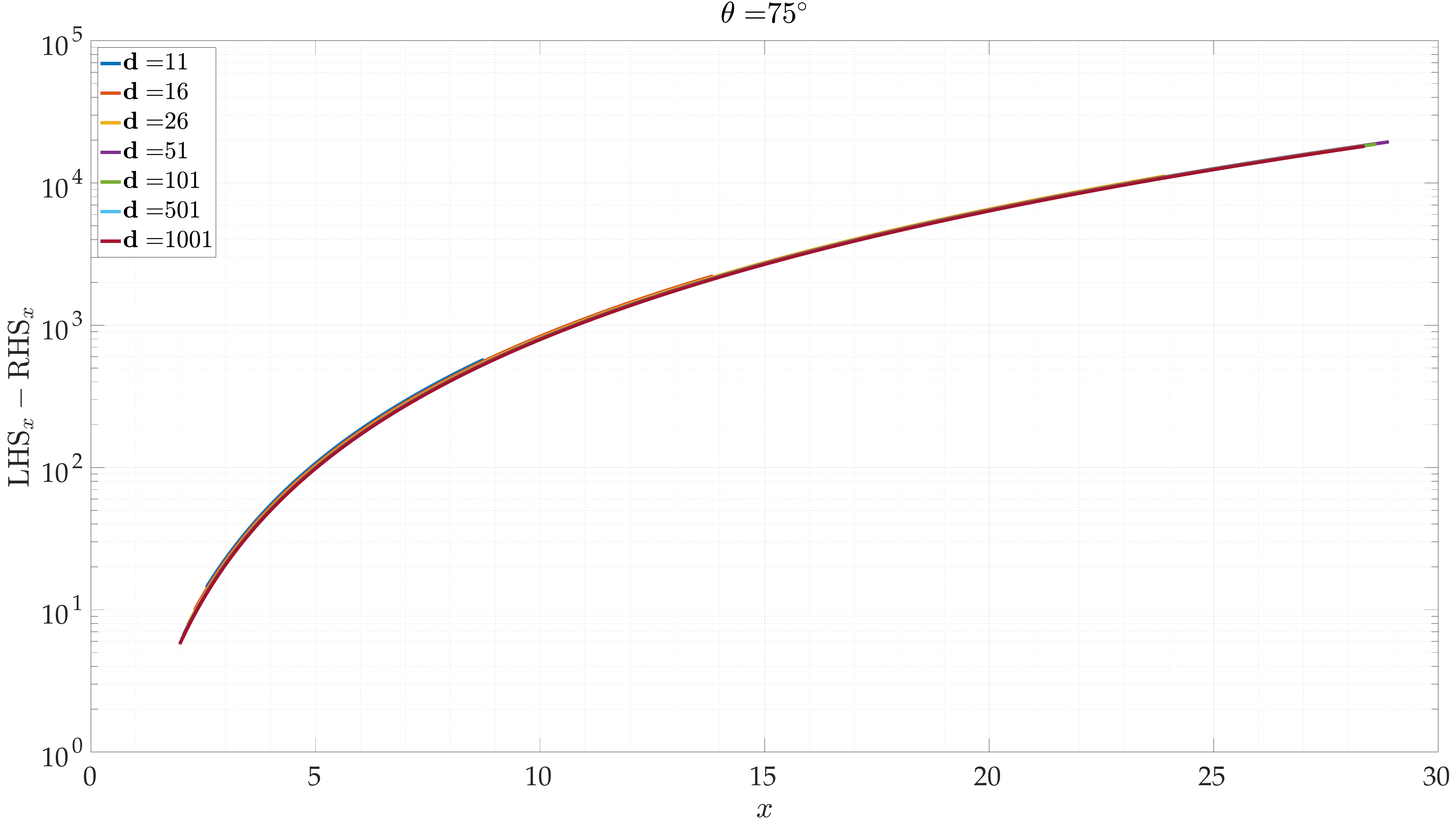}
    \label{fig:3R-1D2B-S-ISO-EQ-NUM-theta-075}
    }
    }
    \caption{Numerical results when evaluating \eqref{eq:3R-1D2B-S-ISO-EQ-g12-g13-Diff-Back} for different values of the vertex angle $ \theta^{\star} $ and gain $ \mathbb{d} $}
    \label{fig:3R-1D2B-S-ISO-EQ-NUM-theta-ALL}
\end{figure*}

Earlier, during the analysis of acute isosceles triangles in Section \ref{sec:Isosceles-Triangle-Analysis-Acute}, we have found that the difference equation \eqref{eq:3R-1D2B-S-ISO-EQ-g12-g13-Diff} contains solutions satisfying Proposition \ref{prop:3R-1D2B-S-ISO-EQ-Conditions-II} for $ \mathbb{d} > 3 $.
Substituting these solutions back to the set of equation \eqref{eq:3R-1D2B-S-ISO-EQ-g12-g13-Set} yields the cubic equation \eqref{eq:3R-1D2B-S-ISO-EQ-g12-g13-Diff-Back} for which we could not determine its sign since positive and negative coefficients are present.
Hence we approach this in a numerical manner. 
We first choose a value for $ \mathbb{d} $ satisfying $ \mathbb{d} > 3 $ ($ \mathbb{d} = \frac{1}{2} R_{\text{Ad}} \cos \theta^{\star} $). 
In the current simulation, we let $ \mathbb{d} \in \begin{bmatrix} 3.1 & 4 & 6 & 11 & 16 & 26 & 51 & 101 & 501 & 1001 \end{bmatrix} $.
Next, we let $ x $ be in the range $ x \in \BR{1, \, 30} $ and compute the corresponding value for $ y $ which solves \eqref{eq:3R-1D2B-S-ISO-EQ-g12-g13-Diff}. 
By applying the constraints found in Proposition \ref{prop:3R-1D2B-S-ISO-EQ-Conditions-II}, we obtain the feasible combinations $ \BR{x_{\text{f}}, \, y_{\text{f}}} $ for the corresponding value of $ \mathbb{d} $. Finally, these combinations $ \BR{x_{\text{f}}, \, y_{\text{f}}} $ are fed back in \eqref{eq:3R-1D2B-S-ISO-EQ-g12-g13-Set}.
We compute the value on the LHS and on the RHS and take the difference between them.
In Fig. \ref{fig:3R-1D2B-S-ISO-EQ-NUM-theta-ALL}, we have plotted the results for $ \theta^{\star} \in \begin{bmatrix} 5^{\degree} & 15^{\degree} & 45^{\degree} & 75^{\degree} \end{bmatrix} $.

From Fig. \ref{fig:3R-1D2B-S-ISO-EQ-NUM-theta-ALL}, we observe that the difference between the LHS and the RHS of \eqref{eq:3R-1D2B-S-ISO-EQ-g12-g13-Set} is positive for the different combinations of $ \mathbb{d} $ and vertex angle $ \theta^{\star} $.
This difference increases for increasing value of $ x $ and also its minimum value increases for increasing value of $ \theta^{\star} $. 
For smaller values of $ \mathbb{d} $, we have a smaller set of $ x $-values in the chosen range which satisfy the constraints in Proposition \ref{prop:3R-1D2B-S-ISO-EQ-Conditions-II}.
From the results of this numerical evaluation, we can conclude that solutions $ \BR{x_{\text{f}}, y_{\text{f}}} $ to the difference equation \eqref{eq:3R-1D2B-S-ISO-EQ-g12-g13-Diff} do not satisfy the set of equations \eqref{eq:3R-1D2B-S-ISO-EQ-g12-g13-Set}.
This also means that the solution set to \eqref{eq:3R-1D2B-S-ISO-EQ-g12-g13-Set} is empty when considering $ x \neq 1 $, $ y \neq 1 $, $ xy \neq 1 $, and $ x \neq q $.
The upper bound for $ R_{\text{Ad}} $ in Lemma \ref{lem:3R-1D2B-S-ISO-EQ-Main-Acute} can be extended to $ \infty $, i.e, we do not need to constrain the gain ration $ R_{\text{Ad}} $.
Given this, we can extend the results for Theorem \ref{thm:3R-1D2B-S-Global-Convergence-Acute} as follows:

\begin{corollary} \label{cor:3R-1D2B-S-Global-Convergence-Acute-V2}
    Consider a team of three robots moving according to \eqref{eq:3R-1D2B-S-p-Dynamics} with $ K_{\text{d}} > 0 $, $ K_{\text{b}} > 0 $, and $ K_{\text{A}} > 0 $. 
    Define the gain ratios $ R_{\text{bd}} = \frac{K_{\text{b}}}{K_{\text{d}}} $ and $ R_{\text{Ad}} = \frac{K_{\text{A}}}{K_{\text{d}}} $, and also $ \widehat{d} = \sqrt{3} \sqrt[3]{\frac{R_{\text{bd}}}{2}} $. 
    Furthermore, let the desired formation shape be an acute isosceles triangle with legs $ \ell > 0 $ and vertex angle $ \theta^{\star} \in \BR{0^{\degree}, \, 90^{\degree}} $. 
    Finally, parametrize the distances $ d_{12} $ and $ d_{13} $ as in \eqref{eq:3R-1D2B-S-ISO-EQ-PAR-XY}.
    Then, starting from all feasible initial configurations, the robots converge to a desired equilibrium configuration $ p_{\text{eq}} \in \mathcal{S}_{p} $ in which all the individual tasks are attained if either $ \BR{K_{\text{d}}, \, K_{\text{b}}} $ is chosen such that $ \ell \leq \widehat{d} $ or if $ \BR{K_{\text{d}}, \, K_{\text{b}}, \, K_{\text{A}}} $ is chosen such that $ \ell > \widehat{d} $ and $ R_{\text{Ad}} \geq \max \CBR{\frac{6}{1 + \cos \theta^{\star}}, \, \frac{2}{1 - \cos \theta^{\star}}} $. 
\end{corollary}

\subsection{Simulation setup for different isosceles triangular formations}
For illustrating the theoretical claims, we consider simulations of isosceles triangles with different values for the legs $ \ell $ and the vertex angle $ \theta^{\star} $.
The gains are taken as $ K_{\text{d}} = 3 $ and $ K_{\text{b}} = 48 $ yielding the threshold distance $ \widehat{d} = 2\sqrt{3} \approx 3.4641 $. 
We let the legs $ \ell $ and vertex angle $ \theta^{\star} $ of the isosceles triangle take values 
\begin{equation} \label{eq:3R-1D2B-S-SIM-ISO-Para}
    \begin{aligned}
        \ell 
            & \in 
        \begin{bmatrix} 
            3 & 6 & 10 
        \end{bmatrix}
        ,
        \\
        \theta^{\star} 
            & \in 
        \begin{bmatrix} 
            5^{\degree} & 10^{\degree} & 30^{\degree} & 60^{\degree} & 90^{\degree} & 120^{\degree} & 150^{\degree} 
        \end{bmatrix}
    \end{aligned}
\end{equation}
while the gain ratio $ R_{\text{Ad}} $ can be chosen from 
\begin{equation} \label{eq:3R-1D2B-S-SIM-RAd-Para}
    \begin{aligned}
        R_{\text{Ad}} 
            & \in 
            \left[
                0.05 \quad 0.1 \quad 0.3 \quad 0.5 \quad 0.7 \quad 0.9 \right. 
                \\
            & \quad \left. 
                \quad  1 \quad 3 \quad 6 \quad 10 \quad 20 \quad 50 \right.
                \\
            & \quad \left.
                \quad \frac{2}{1 - \cos \theta^{\star}} \quad \frac{2}{1 + \cos \theta^{\star}} \quad \frac{6}{1 + \cos \theta^{\star}}
            \right]
            .
    \end{aligned}
\end{equation}
The initial positions of the robots are in the square $ \left[-100, \, 100 \right]^{2} $ and we consider simulations from $ 5000 $ starting positions for the team of robots. 

\subsection{Simulation results for different isosceles triangular formations}
\begin{figure*}[!tb]
    \centering
    {
    \subfigure[Isosceles Triangle]
    {
    \includegraphics[width=0.475\textwidth]{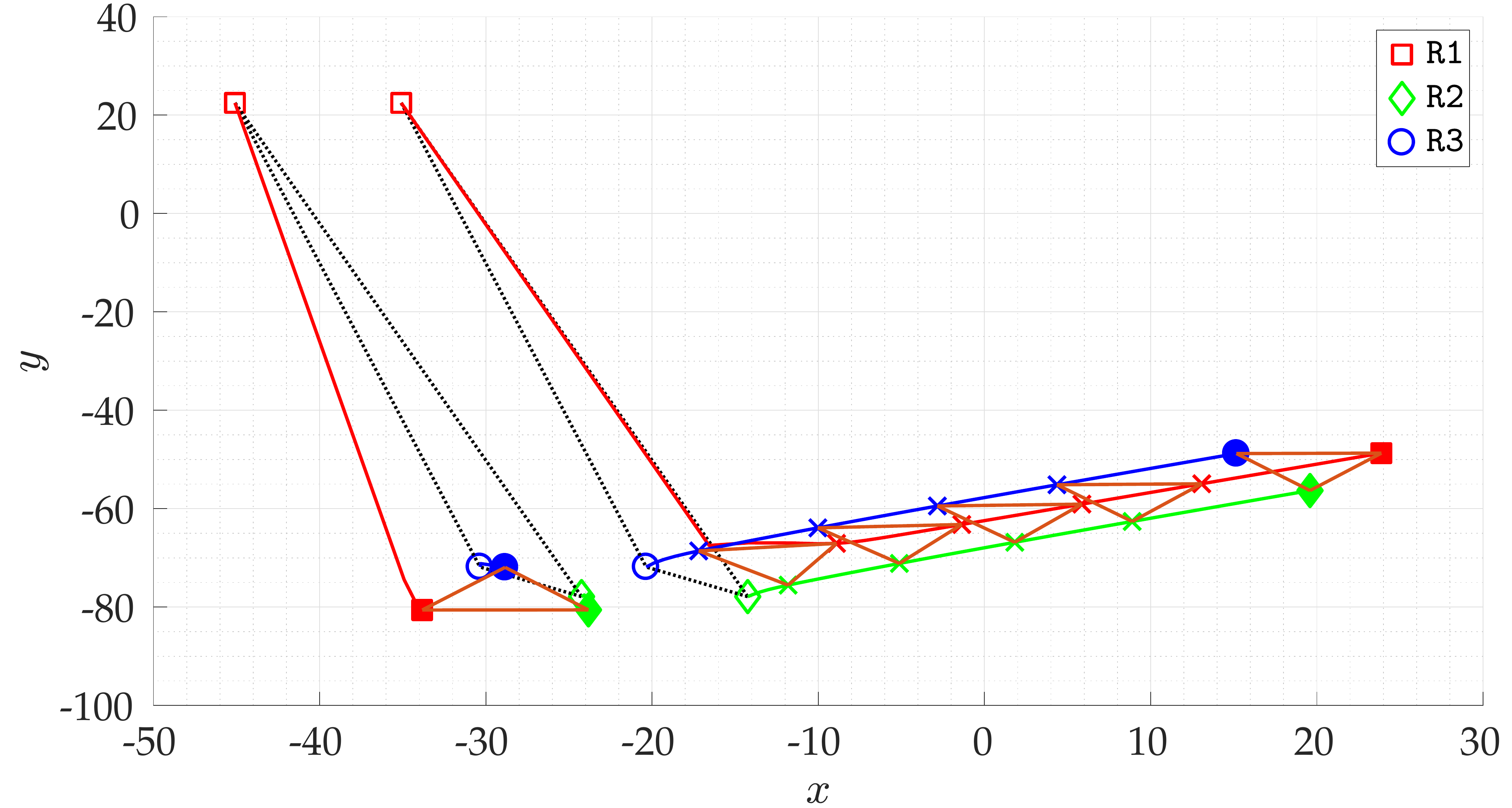}
    \label{fig:3R-1D2B-S-ISO-Conv}
    }
    \hfill
    \subfigure[General Triangle]
    {\includegraphics[width=0.475\textwidth]{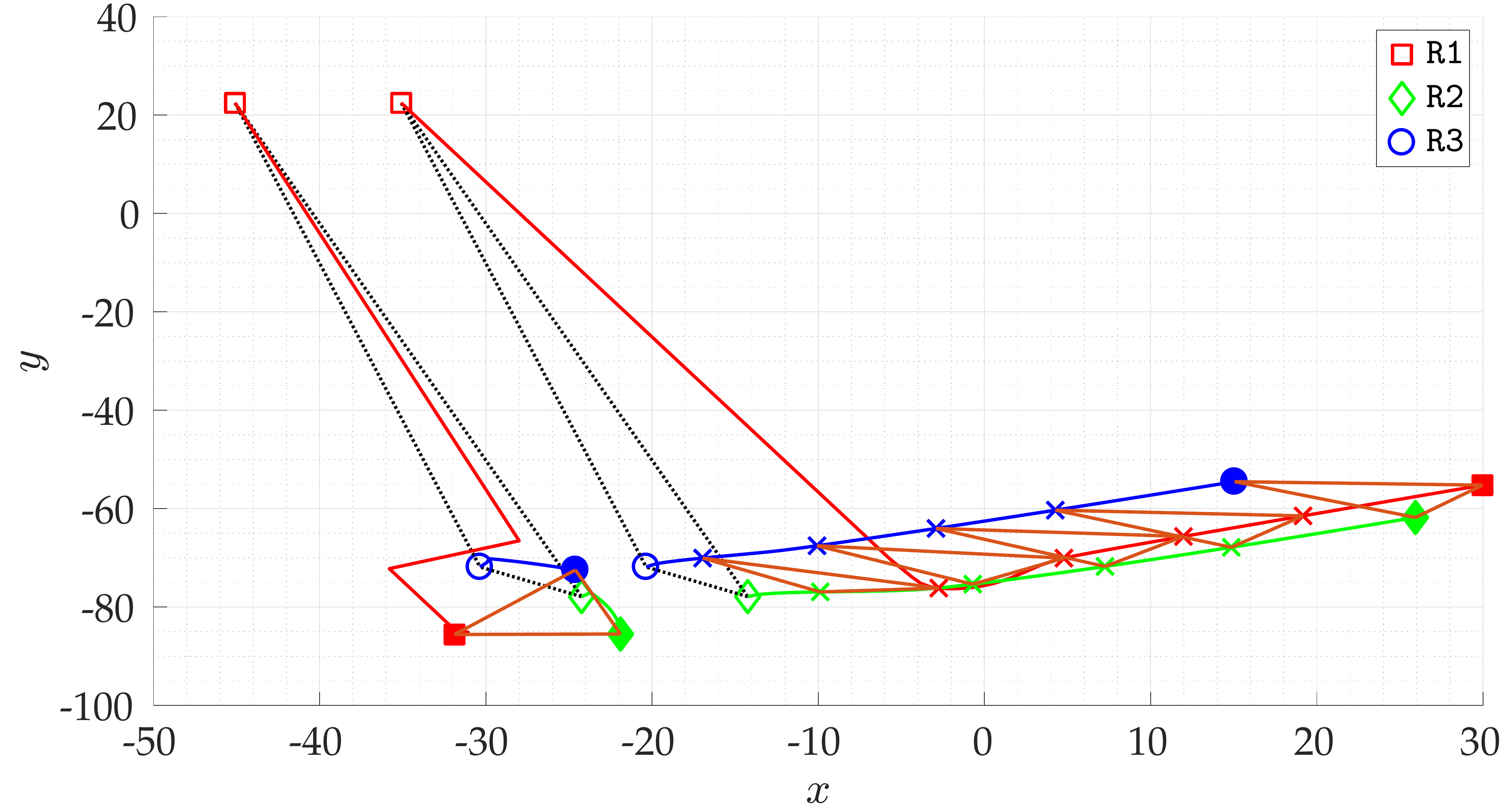}
    \label{fig:3R-1D2B-S-GEN-Conv}
    }
    }
    \caption{Robot trajectories for the (\textbf{1D2B}) setup; \textbf{Left}: Desired shape is an equilateral triangle with leg $ \ell = 10 $. For $ R_{\text{Ad}} = 0.5 $, the trajectories converge to an incorrect moving configuration while for $ R_{\text{Ad}} = 1 $, the trajectories converge to the desired equilateral triangle. \textbf{Right}: Desired shape is a general triangle with $ d_{12}^{\star} = 10 $, $ d_{13}^{\star} = 15 $, and $ \theta_{213} = 60^{\degree} $. The incorrect moving configuration occurs when $ R_{\text{Ad}} = 0.2 $ while convergence to the desired shape is obtained when $ R_{\text{Ad}} = 1 $.
    }
    \label{fig:3R-1D2B-S-ISO-GEN-CONV}
\end{figure*}

Here we present the results from the numerical set up. 
For the isosceles triangle with legs $ \ell = 3 < \widehat{d} $, we obtain that starting from all the considered initial positions, the robots converge to a desired equilibrium configuration in $ \mathcal{S}_{p} $. 
For isosceles triangles with legs $ \ell = 6 $ and $ \ell = 10 $, we observe that for small values of $ R_{\text{Ad}} $ less than $ 1 $, we have convergence to moving configurations. 
For $ \ell = 6 $, this is $ R_{\text{Ad}} \leq 0.5 $ while for $ \ell = 10 $, we have $ R_{\text{Ad}} \leq 0.7 $. 
To be safe, we can infer from the current results that starting from $ R_{\text{Ad}} = 1 $, we only have convergence to a desired equilibrium. 
This value is smaller than the lower bound that we have obtained in Lemmas \ref{lem:3R-1D2B-S-ISO-MV-Main-Acute} and \ref{lem:3R-1D2B-S-ISO-MV-Main-Right-Obtuse}.
To better illustrate the convergence observation, we plot the results of two simulations in Fig. \ref{fig:3R-1D2B-S-ISO-Conv}.
The desired shape is an equilateral triangle with legs $ \ell = 10 $. 
We start from the same initial position. In order to show the simulation results clearly, we plot them side-by-side by shifting the trajectories horizontally. 
For $ R_{\text{Ad}} = 0.5 $, we observe that the robots converge to a moving configuration while choosing $ R_{\text{Ad}} = 1 $ results in convergence to the desired formation shape.

In order to observe whether for larger values of $ \ell $ we also have this result, we consider taking $ \theta^{\star} = 60^{\degree} $; this corresponds to an equilateral triangle. 
Now, we let $ \ell $ and $ R_{\text{Ad}} $ take values 
\begin{equation} \label{eq:3R-1D2B-S-SIM-ISO-Para-V2}
    \begin{aligned}
        \ell 
            & \in 
            \left[
                3 \quad 4 \quad 6 \quad 8 \quad 10 \quad 15 
                \quad 20 \quad 25 \quad 50 \quad 75 \quad 100
            \right]
        ,
        \\
        R_{\text{Ad}}
            & \in 
        \begin{bmatrix} 
            0.25 & 0.5 & 0.75 & 1 & 2 & 4
        \end{bmatrix}
        .
    \end{aligned}
\end{equation}
We observe that convergence to moving configurations occur when $ R_{\text{Ad}} \leq 0.75 $ while starting from $ R_{\text{Ad}} = 1 $, all initial configurations evolve to a desired configuration in the set $ \mathcal{S}_{p} $. 
From Lemma \ref{lem:3R-1D2B-S-ISO-MV-Main-Acute}, we have that the theoretical lower bound is $ R_{\text{Ad}} = 4 $ while numerically $ R_{\text{Ad}} = 1 $ suffices. 

From these numerical results, we can say that the bounds for $ R_{\text{Ad}} $ obtained during the theoretical analyses are conservative and that a value $ R_{\text{Ad}} \geq 1 $ suffices to prevent the occurrence of moving configurations for isosceles triangles. 

\subsection{Extension to general triangles}
So far, we have obtained results for the analysis of the (\textbf{1D2B}) setup with a signed area constraint when the desired formation shape is an isosceles triangle. 
To demonstrate that the proposed closed-loop formation system \eqref{eq:3R-1D2B-S-p-Dynamics} may also work for general triangles, we have carried out some simulations.
In Fig. \ref{fig:3R-1D2B-S-GEN-Conv}, we plot the result of two simulations.
The desired formation shape is a general triangle with lengths $ d_{12}^{\star} = 10 $ and $ d_{13}^{\star} = 15 $ and angle $ \theta_{213} = 60^{\degree} $. 
The moving configuration is obtained when $ R_{\text{Ad}} = 0.2 $ while we observe convergence to the desired shape when $ R_{\text{Ad}} = 1 $. 
This illustrates that by a proper tuning of the gains, we could also have convergence results to $ \mathcal{S}_{p} $ for general triangles.
Furthermore, we notice that when one of the desired lengths is less than $ \widehat{d} $, we always have convergence to the desired formation shape. 
This is in accordance with our earlier work \cite{Chan2020}. 


\section{Conclusions \& Future Work} \label{sec:Conclusions}
In this paper, we provided a comprehensive analysis for the formation shape control problem involving a team of three robots partitioned into one distance and two bearing robots. 
We let the distance robot also maintained a signed area constraint next to the existing distance constraints considered in our previous work \cite{Chan2020}, and studied the effect of this new constraint for the class of isosceles triangles.
We showed theoretically and using numerical simulations that the existing equilibrium configurations were maintained and no other undesired equilibrium configurations were introduced by the addition of the signed area control term. 
Moreover, we derived sufficient conditions on the gain ratio $ R_{\text{Ad}} $ for preventing moving configurations to occur when the leg $ \ell $ of the triangle is larger than a threshold distance $ \widehat{d} $.
As a result, convergence results to the desired set $ \mathcal{S}_{p} $ were established for arbitrary isosceles triangular formations. 
Numerical results indicated that a lower bound of $ R_{\text{Ad}} = 1 $ suffices for preventing convergence to moving configurations while the theoretical analyses resulted in a more restrictive value that depends on the vertex angle $ \theta^{\star} $.
Furthermore, simulations showed that the proposed strategy could also work for general triangles. 
The formal analysis of general triangles is the subject of future work. 


\bibliographystyle{abbrv} 
\bibliography{Literature-root}

\begin{thebibliography}{10}

\bibitem{Anderson2017a}
B.~D.~O. Anderson, Z.~Sun, T.~Sugie, S.-I. Azuma, and K.~Sakurama.
\newblock Distance-based rigid formation control with signed area constraints.
\newblock In {\em 2017 56th {IEEE} Conference on Decision and Control ({CDC})},
  Dec. 2017.

\bibitem{anderson2007control}
B.~D.~O. Anderson, C.~Yu, S.~Dasgupta, and A.~S. Morse.
\newblock Control of a three-coleader formation in the plane.
\newblock {\em Systems \& Control Letters}, 56(9):573--578, 2007.

\bibitem{Anderson2008}
B.~D.~O. {Anderson}, C.~{Yu}, B.~{Fidan}, and J.~M. {Hendrickx}.
\newblock Rigid graph control architectures for autonomous formations.
\newblock {\em IEEE Control Systems Magazine}, 28(6):48--63, Dec. 2008.

\bibitem{Bishop2013}
A.~N. Bishop, T.~H. Summers, and B.~D.~O. Anderson.
\newblock {Stabilization of stiff formations with a mix of direction and
  distance constraints}.
\newblock In {\em 2013 {IEEE} International Conference on Control Applications
  ({CCA})}, Aug. 2013.

\bibitem{Cao2007}
M.~Cao, A.~S. Morse, C.~Yu, B.~D.~O. Anderson, and S.~Dasgupta.
\newblock Controlling a triangular formation of mobile autonomous agents.
\newblock In {\em 2007 46th {IEEE} Conference on Decision and Control}, 2007.

\bibitem{Cao2008}
M.~Cao, C.~Yu, A.~Morse, B.~D.~O. Anderson, and S.~Dasgupta.
\newblock Generalized controller for directed triangle formations.
\newblock {\em {IFAC} Proceedings Volumes}, 41(2):6590--6595, 2008.

\bibitem{Cao2019a}
Y.~Cao, Z.~Sun, B.~D.~O. Anderson, and T.~Sugie.
\newblock Almost global convergence for distance- and area-constrained
  hierarchical formations without reflection.
\newblock In {\em 2019 15th {IEEE} International Conference on Control and
  Automation ({ICCA})}, July 2019.

\bibitem{Chan2020}
N.~P.~K. Chan, B.~Jayawardhana, and H.~G. de~Marina.
\newblock Stability analysis of gradient-based distributed formation control
  with heterogeneous sensing mechanism: Two and three robot case.
\newblock {\em https://arxiv.org/abs/2010.10559}, 2020.

\bibitem{Chen2020}
L.~Chen, M.~Cao, and C.~Li.
\newblock Angle rigidity and its usage to stabilize multi-agent formations in
  2{D}.
\newblock {\em {IEEE} Transactions on Automatic Control}, pages 1--1, 2020.

\bibitem{Marina2017}
H.~G. de~Marina, Z.~Sun, M.~Cao, and B.~D.~O. Anderson.
\newblock Controlling a triangular flexible formation of autonomous agents.
\newblock {\em {IFAC}-{PapersOnLine}}, 50(1):594--600, July 2017.

\bibitem{Kwon2019a}
S.-H. Kwon and H.-S. Ahn.
\newblock {Generalized Rigidity and Stability Analysis on Formation Control
  Systems with Multiple Agents}.
\newblock In {\em 2019 18th European Control Conference ({ECC})}, June 2019.

\bibitem{Kwon2018}
S.-H. Kwon, M.~H. Trinh, K.-H. Oh, S.~Zhao, and H.-S. Ahn.
\newblock {Infinitesimal Weak Rigidity and Stability Analysis on Three-Agent
  Formations}.
\newblock In {\em 2018 57th Conference of the Society of Instrument and Control
  Engineers of Japan ({SICE})}, Sept. 2018.

\bibitem{liu2014controlling}
H.~Liu, H.~G. de~Marina, and M.~Cao.
\newblock Controlling triangular formations of autonomous agents in finite time
  using coarse measurements.
\newblock In {\em 2014 IEEE International Conference on Robotics and Automation
  (ICRA)}, pages 3601--3606, 2014.

\bibitem{7039453}
S.~Mou, A.~S. Morse, M.~A. Belabbas, and B.~D.~O. Anderson.
\newblock Undirected rigid formations are problematic.
\newblock In {\em 2014 53rd {IEEE} Conference on Decision and Control}, pages
  637--642, Dec. 2014.

\bibitem{Oh2015}
K.-K. Oh, M.-C. Park, and H.-S. Ahn.
\newblock A survey of multi-agent formation control.
\newblock {\em Automatica}, 53:424--440, Mar. 2015.

\bibitem{Sugie2018}
T.~Sugie, B.~D.~O. Anderson, Z.~Sun, and H.~Dong.
\newblock On a hierarchical control strategy for multi-agent formation without
  reflection.
\newblock In {\em 2018 57th {IEEE} Conference on Decision and Control ({CDC})},
  Dec. 2018.

\bibitem{Sugie2020}
T.~Sugie, F.~Tong, B.~D.~O. Anderson, and Z.~Sun.
\newblock On global convergence of area-constrained formations of hierarchical
  multi-agent systems.
\newblock {\em To be presented at 2020 59th {IEEE} Conference on Decision and
  Control; https://arxiv.org/abs/2009.03048}, 2020.

\bibitem{Zhao2019}
S.~Zhao and D.~Zelazo.
\newblock {Bearing Rigidity Theory and Its Applications for Control and
  Estimation of Network Systems: Life Beyond Distance Rigidity}.
\newblock {\em {IEEE} Control Systems Magazine}, 39(2):66--83, Apr. 2019.

\end{thebibliography}

\end{document}